\pdfoutput=1

\documentclass[11pt]{article}

\usepackage[final]{acl}

\usepackage{times}
\usepackage{latexsym}

\usepackage[T1]{fontenc}

\usepackage[utf8]{inputenc}

\usepackage{microtype}

\usepackage{inconsolata}

\usepackage{hyperref}
\hypersetup{
colorlinks   = true,
citecolor    = blue,
urlcolor    =  blue
}
\usepackage{url}
\usepackage{cleveref}

\usepackage{float}
\usepackage{dblfloatfix} 

\usepackage{subcaption}
\usepackage[
singlelinecheck=false 
]{caption}

\usepackage{hhline}
\usepackage{multirow}
\usepackage{tabularx}
\newcolumntype{Y}{>{\centering\arraybackslash}X}
\newcolumntype{Z}{>{\raggedleft\arraybackslash}X}
\usepackage{makecell}
\usepackage{dcolumn}
\newcolumntype{d}[1]{D{.}{.}{#1}}
\newcolumntype{t}[1]{D{:}{:}{#1}}

\usepackage{pgfplots} 
\usepackage{colortbl}

\crefformat{footnote}{#2\footnotemark[#1]#3}
\newcommand{\footURL}[1]{\footnote{\url{#1}}}

\makeatletter
\newcommand\footnoteref[1]{\protected@xdef\@thefnmark{\ref{#1}}\@footnotemark}
\makeatother

\usepackage{ragged2e}
\usepackage{pdflscape}
\usepackage{svg}
\usepackage{xcolor}
\usepackage{soul}

\usepackage{lipsum} 

\newcommand*{\affmark}[1][*]{\textsuperscript{#1}}

\title{\textit{Shoulders of Giants:} A Look at the Degree and Utility of Openness in NLP Research}

\author{Surangika Ranathunga\affmark[1], Nisansa de Silva\affmark[2], Dilith Jayakody\affmark[2], Aloka Fernando\affmark[2]\\
  \affmark[1]School of Mathematical and Computational Sciences, Massey University, New Zealand \\
   \texttt{s.ranathunga@massey.ac.nz} \\
 \affmark[2]Dept. of Computer Science \& Engineering, University of Moratuwa,10400, Sri Lanka \\
  \texttt{\{NisansaDdS,dilith.18,alokaf\}@cse.mrt.ac.lk}  
}

\begin{document}
\maketitle
\begin{abstract}
We analysed a sample of NLP research papers archived in ACL Anthology as an attempt to quantify the degree of openness and the benefit of such an open culture in the NLP community. We observe that papers published in different NLP venues show different patterns related to artefact reuse. We also note that more than 30\% of the papers we analysed do not release their artefacts publicly, despite promising to do so. Further, we observe a wide language-wise disparity in publicly available NLP-related artefacts. 
\end{abstract}

\section{Introduction}
The advancement of the Computer Science research field heavily depends on publicly available code, software, and tools. Its sub-fields Machine Learning and Natural Language Processing (NLP) have the additional requirement of datasets - to train and evaluate computational models. Lack of access to these research artefacts has been identified as a major reason for the difficulty in reproducing works of others~\cite{pineau2021improving}.  The data requirement is particularly challenging in NLP - a dataset available for one language usually cannot be used in the context of another language\footnote{Other than in techniques such as multi-tasking and intermediate-task fine-tuning.}.

Therefore, the NLP community is highly encouraged to make their research artefacts publicly available. However, as far as we are aware, there is no quantifiable evidence on (1) the degree of openness in the NLP community or (2) the benefit of openness to the community. Since  \textit{``what we do not measure, we cannot
improve''}~\cite{rungta-etal-2022-geographic}, in this paper, we quantify both these aspects. To this end, we semi-automatically analyse a sample of NLP research papers published in ACL Anthology (AA) and corpora/ Language Models (LMs) released in {Hugging Face}\footnote{\url{https://huggingface.co/}}, and answer the following questions:

\begin{enumerate}
    \item To what degree has the NLP research community been able to reuse open-source artefacts (data, code, LMs) in their research?
    \item How much has the community freely shared the artefacts produced by their research?
\end{enumerate}

To answer the first question, we record the number of papers that reuse the artefacts released by past research. Since there is a language-wise disparity in NLP research~\cite{joshi-etal-2020-state, ranathunga-de-silva-2022-languages}, this analysis is conducted while separating low- and high-resource languages.

To answer the second question, we record the papers that indicate they would release the newly produced artefacts. We also record whether they have provided a repository URL. We do further analysis to find out whether these repositories have the artefacts they are supposed to have. Finally, we record the number of datasets and LMs available for different language classes on Hugging Face.

\begin{figure}[!htb]
    \captionsetup[subfigure]{justification=centering}
    \begin{subfigure}{0.48\textwidth}  
        \centering
        \includegraphics[width=\linewidth]{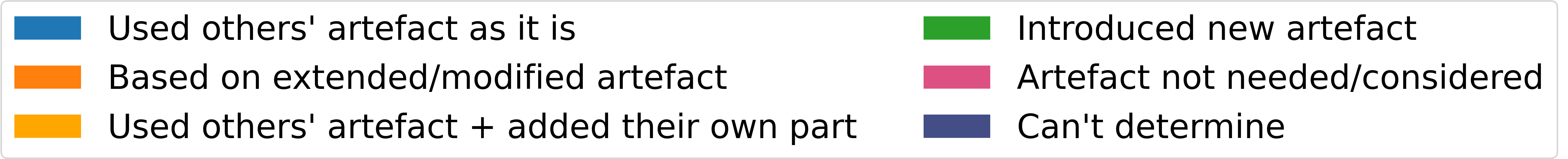}
    \end{subfigure}
    
    \centering
    \begin{subfigure}{0.15\textwidth} 
        \centering
        \includegraphics[width=\linewidth]{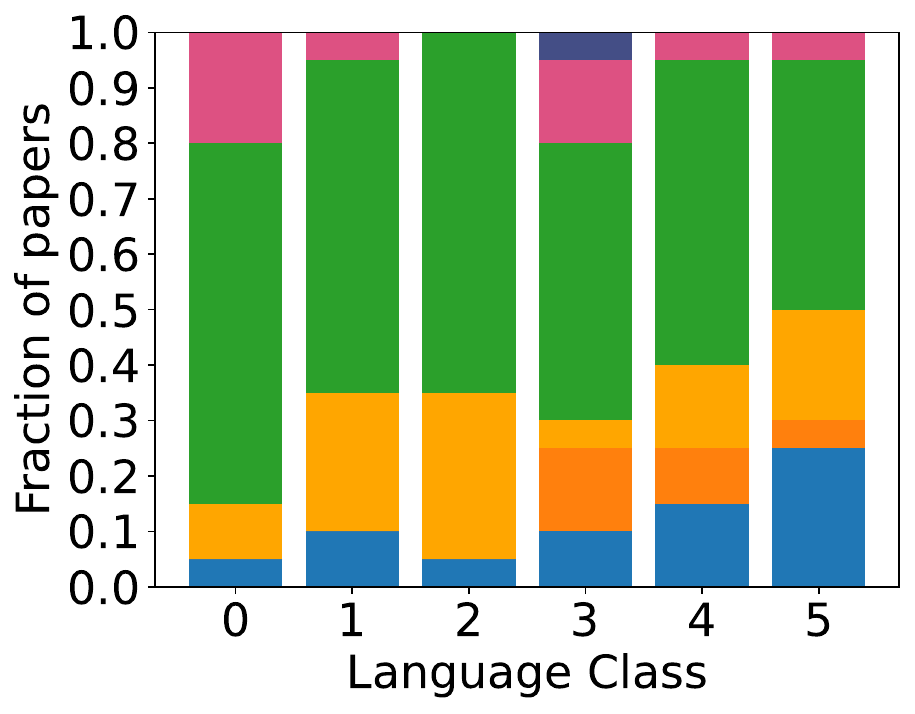}
        \caption{Data - LREC}
        \label{fig:LRECdata}
    \end{subfigure}
    \hfill
    \begin{subfigure}{0.15\textwidth}
        \centering
        \includegraphics[width=\linewidth]{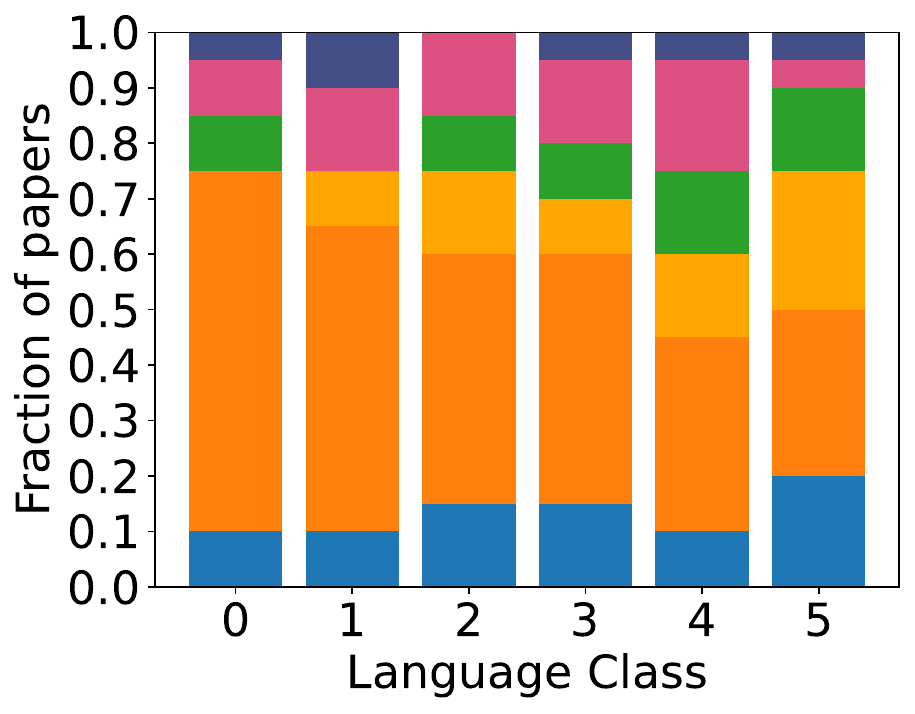}
        \caption{Code - LREC}
        \label{fig:LRECcode}
    \end{subfigure}
    \hfill
    \begin{subfigure}{0.15\textwidth}
        \centering
        \includegraphics[width=\linewidth]{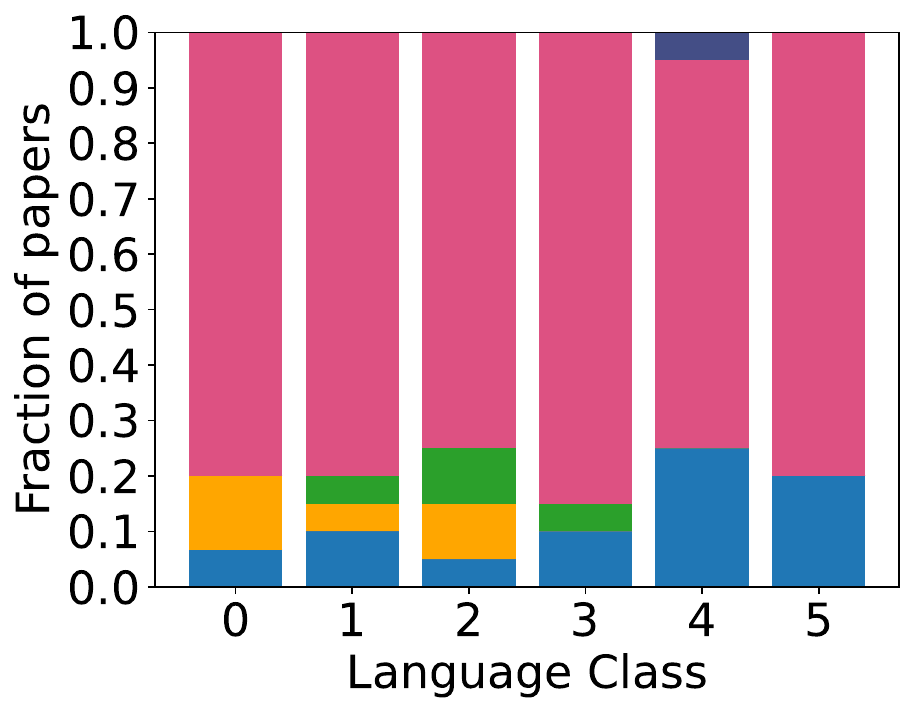}
        \caption{LMs - LREC}
        \label{fig:LRECmodel}
    \end{subfigure}

    \begin{subfigure}{0.15\textwidth}  
        \centering
        \includegraphics[width=\linewidth]{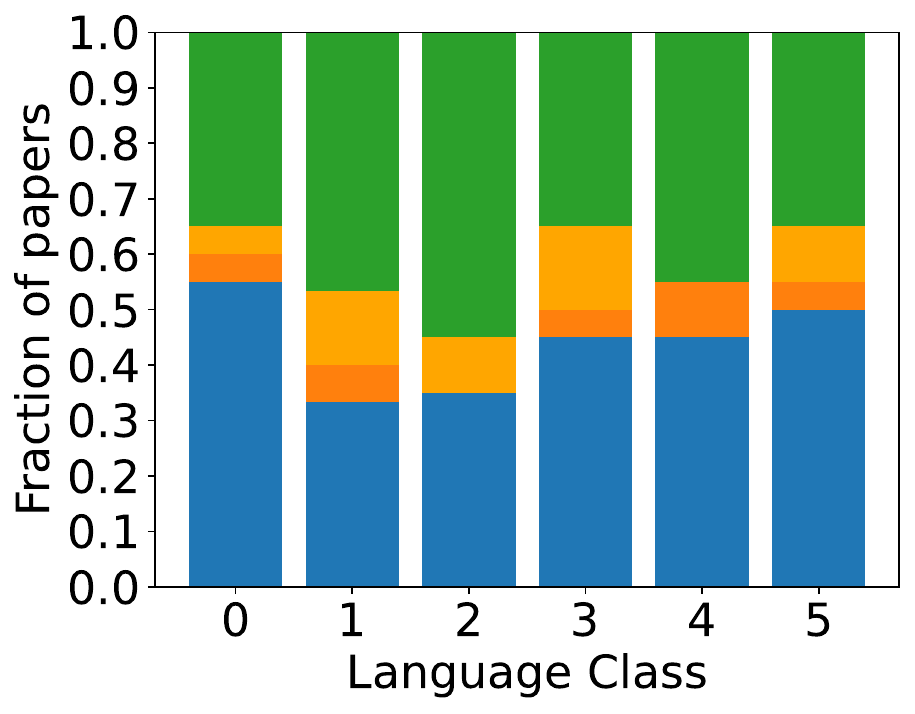}
        \caption{Data - Main}
        \label{fig:data}
    \end{subfigure}
    \hfill
    \begin{subfigure}{0.15\textwidth}
        \centering
        \includegraphics[width=\linewidth]{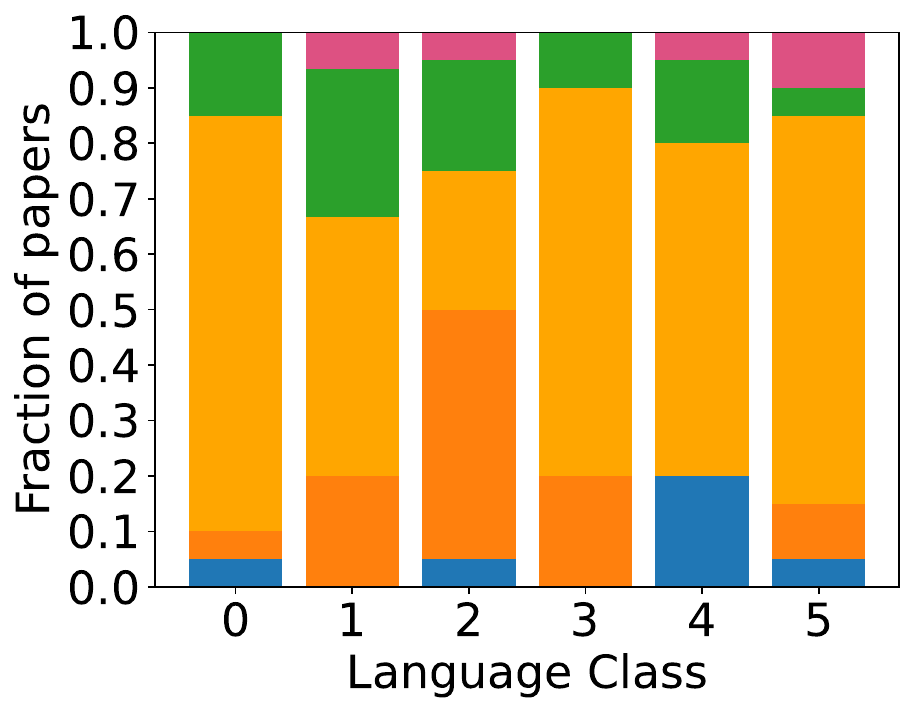}
        \caption{Code - Main}
        \label{fig:code}
    \end{subfigure}
    \hfill
    \begin{subfigure}{0.15\textwidth}
        \centering
        \includegraphics[width=\linewidth]{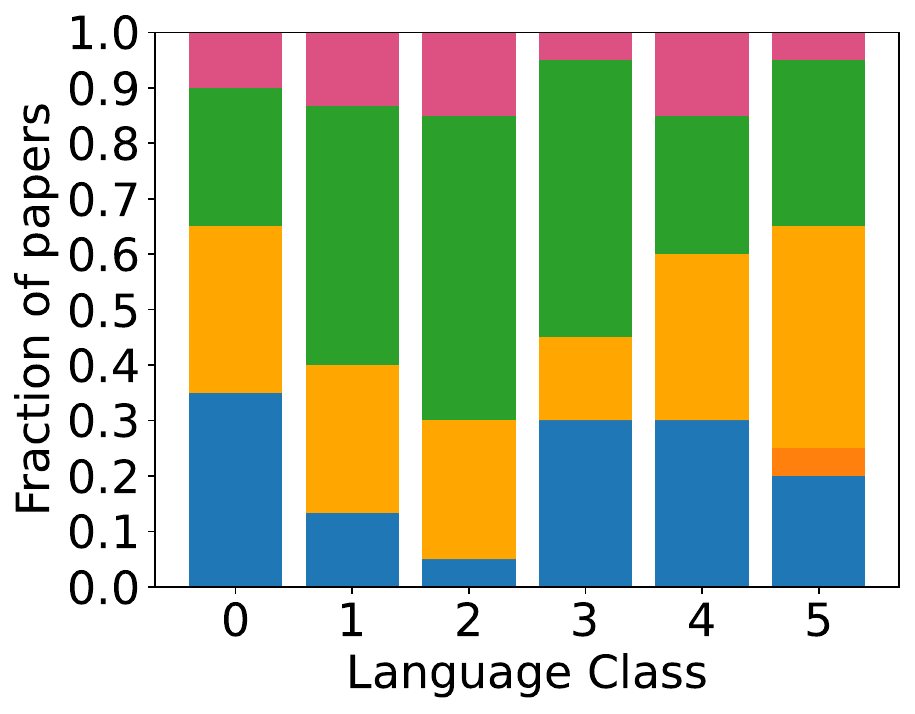}
        \caption{LMs - Main}
        \label{fig:model}
    \end{subfigure}   

    \begin{subfigure}{0.15\textwidth}  
        \centering
        \includegraphics[width=\linewidth]{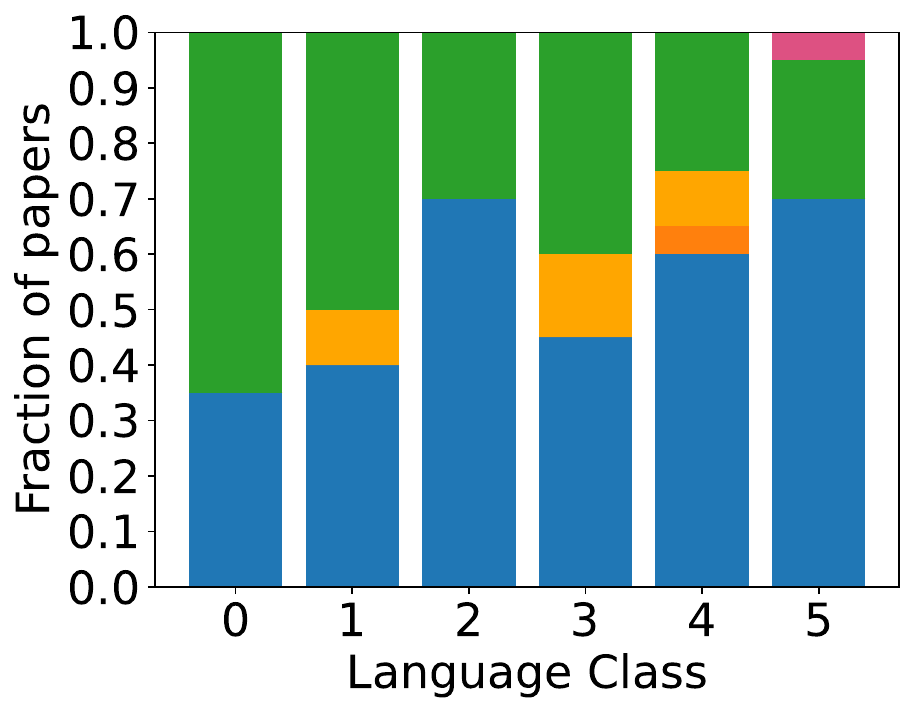}
        \caption{Data - Other}
        \label{fig:data}
    \end{subfigure}
    \hfill
    \begin{subfigure}{0.15\textwidth}
        \centering
        \includegraphics[width=\linewidth]{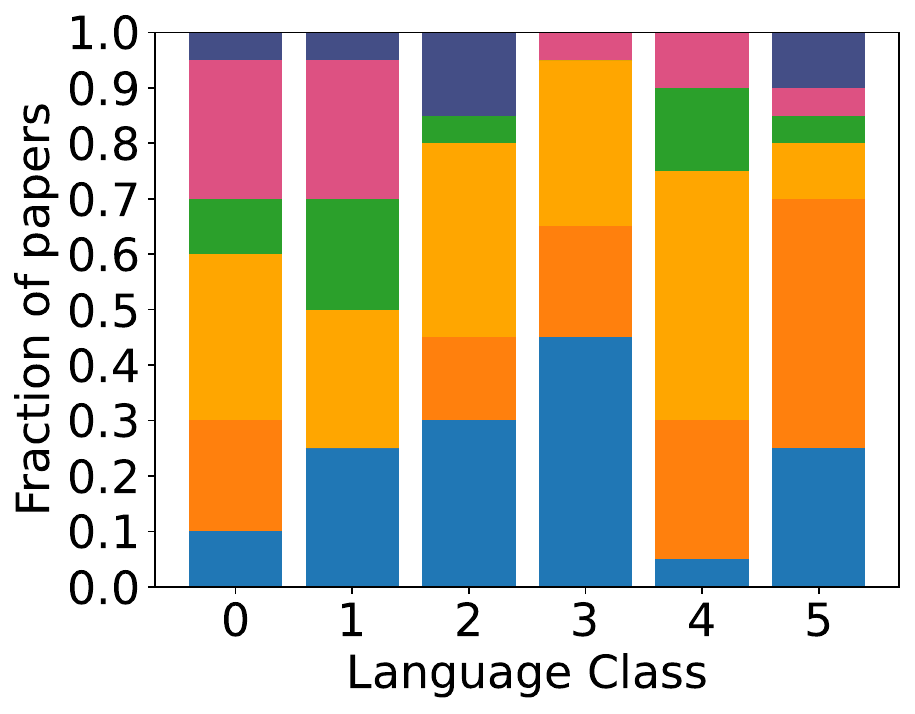}
        \caption{Code - Other}
        \label{fig:code}
    \end{subfigure}
    \hfill
    \begin{subfigure}{0.15\textwidth}
        \centering
        \includegraphics[width=\linewidth]{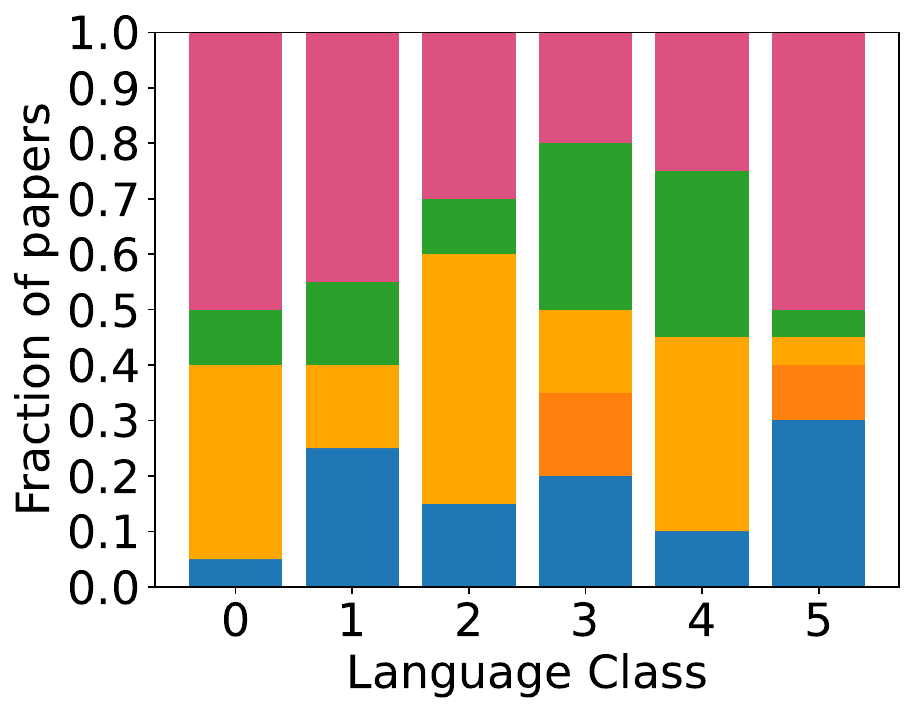}
        \caption{LMs - Other}
        \label{fig:model}
    \end{subfigure}  
    \centering
    \caption{Artefact (Data, Code, and LMs) creation, extension, and reuse across PVs.}
    \label{fig:resource-usage-ACL}
\end{figure}

We observe that papers published in different venues show different patterns in artefact reuse. We also observe that a worrying percentage of papers that produced an artefact have not publicly released those artefacts. To a lesser degree, broken repository links and empty resource repositories were also noted. Finally, it is noted that the language-wise disparity in LM/data availability~\cite{joshi-etal-2020-state, ranathunga-de-silva-2022-languages, khanuja-etal-2023-evaluating} is still staggering.

\section{Data Extraction}
\label{sec:methodology}

\begin{figure*}[!hbt]
    \captionsetup[subfigure]{justification=centering}
    \centering
    \begin{subfigure}{0.32\textwidth} 
        \centering
        \includegraphics[width=\linewidth]{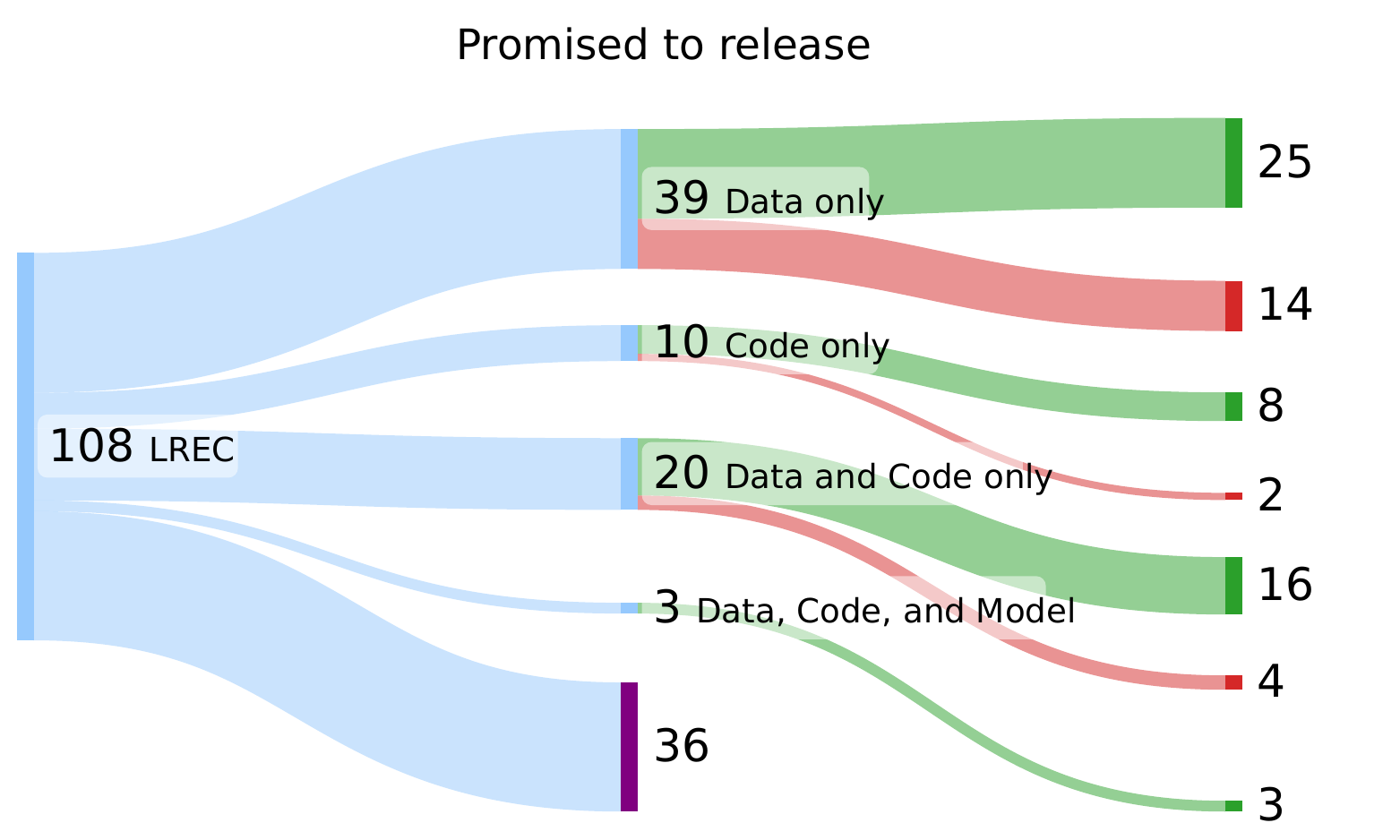}
        \caption{LREC}
        \label{fig:LRECdata}
    \end{subfigure}
    \hfill
    \begin{subfigure}{0.32\textwidth}
        \centering
        \includegraphics[width=\linewidth]{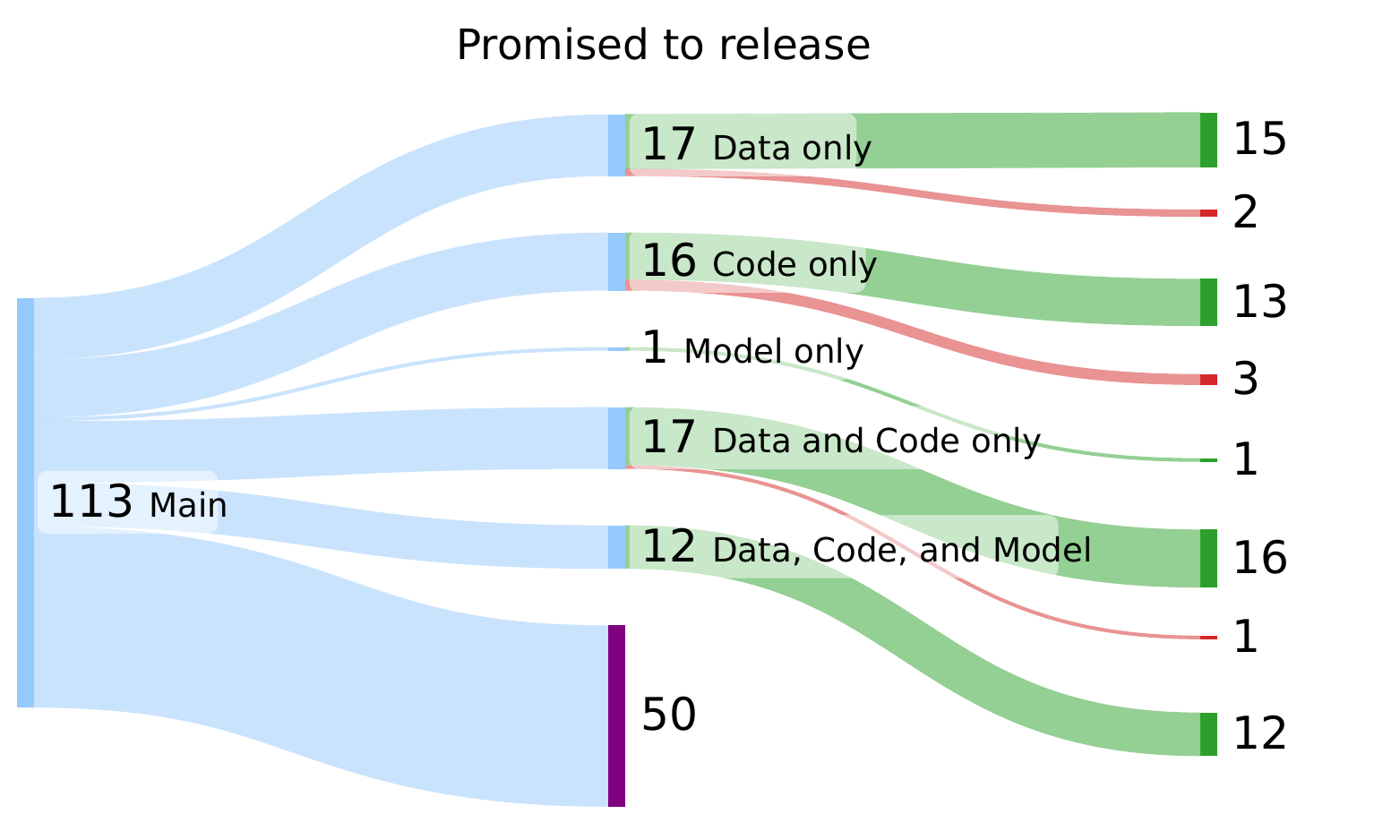}
        \caption{Main}
        \label{fig:LRECcode}
    \end{subfigure}
    \hfill
    \begin{subfigure}{0.32\textwidth}
        \centering
        \includegraphics[width=\linewidth]{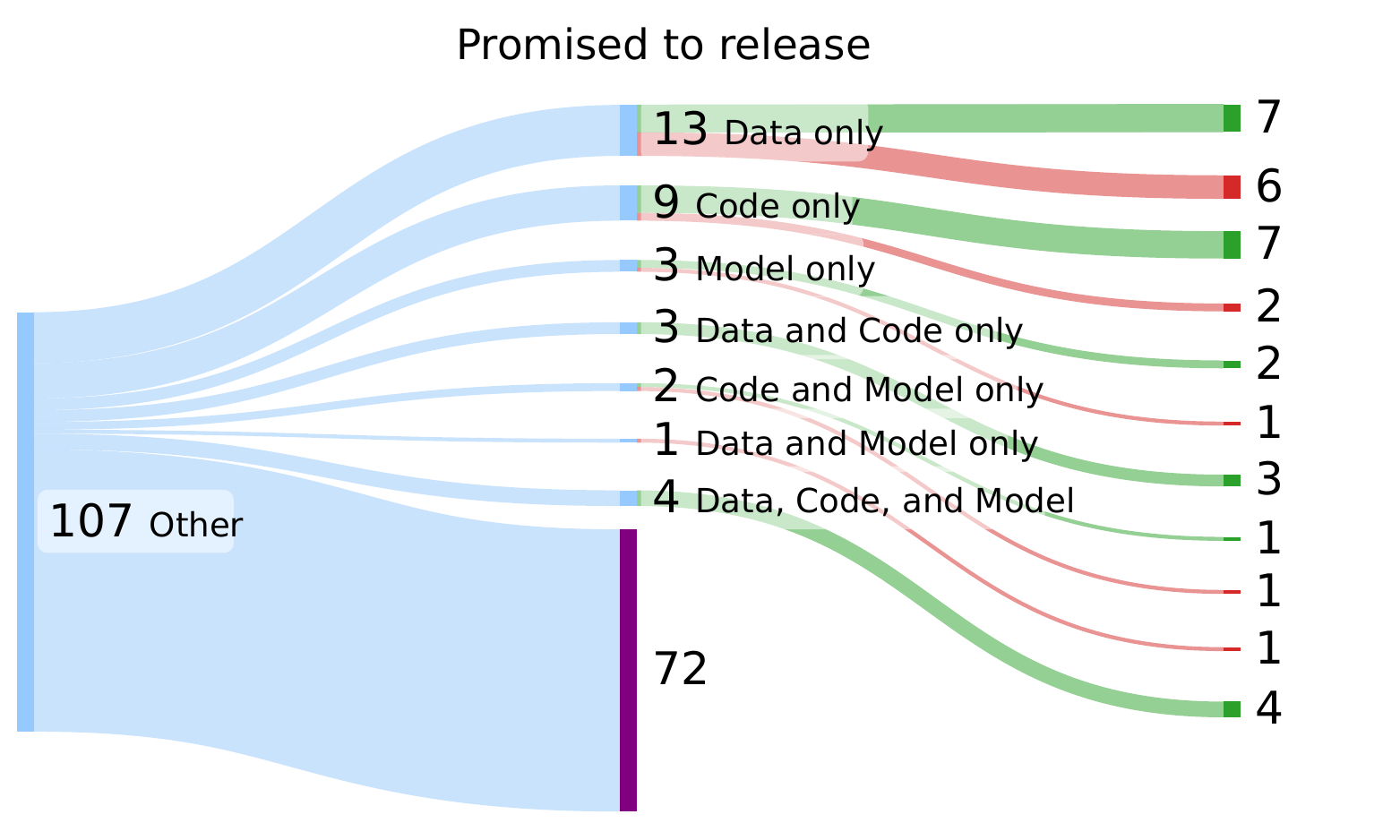}
        \caption{Other}
        \label{fig:LRECmodel}
    \end{subfigure}
    \centering
    \caption{Artefact releasing promise vs artefact link availability across PVs. Green - Artefact Released, Red - Claimed to release the relevant artefact but no link given, Purple - No promise was given to release any artefact.}
    \label{fig:artefact-release-promises-ACL}
\end{figure*}

\begin{figure*}[!hbt]
    \captionsetup[subfigure]{justification=centering}
    \begin{minipage}{\textwidth}
        \centering
        \includegraphics[width=0.4\textwidth]{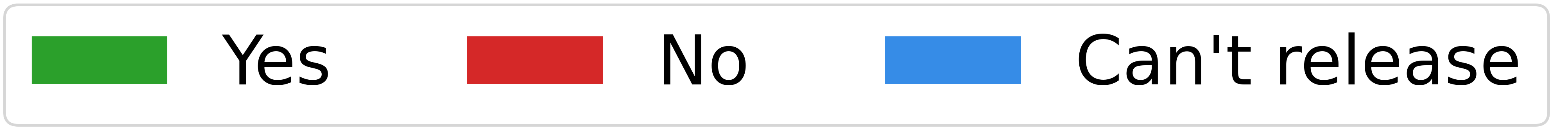}
    \end{minipage}
    \centering
    \begin{subfigure}{0.24\textwidth}  
        \centering
        \includegraphics[width=\linewidth]{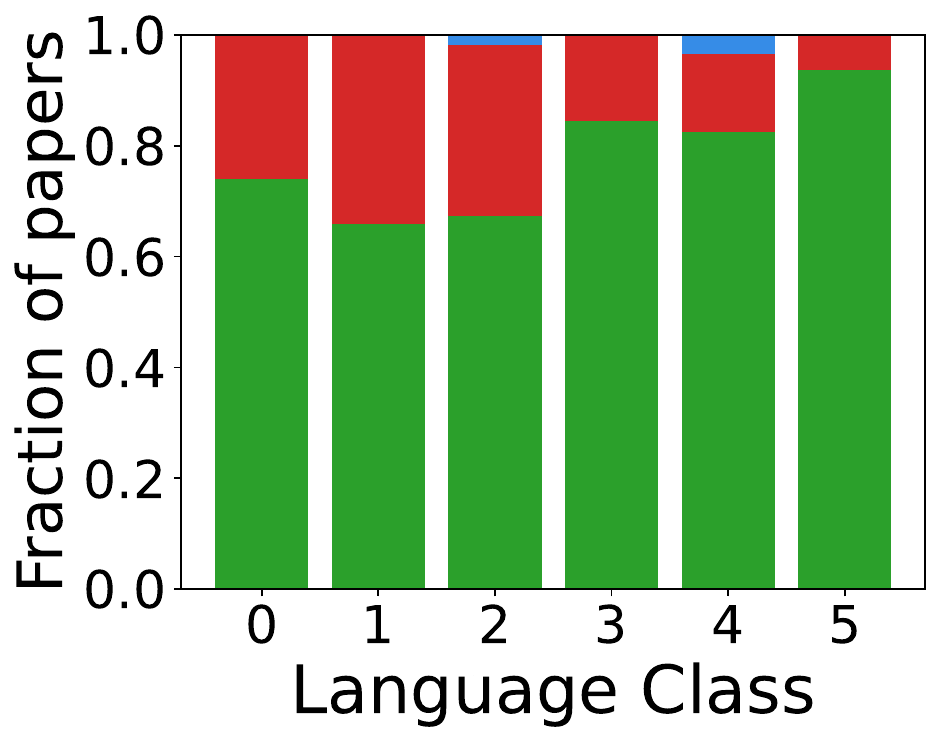}
        \caption{Availability of Data}
        \label{fig:data}
    \end{subfigure}
    \hfill
    \begin{subfigure}{0.24\textwidth}
        \centering
        \includegraphics[width=\linewidth]{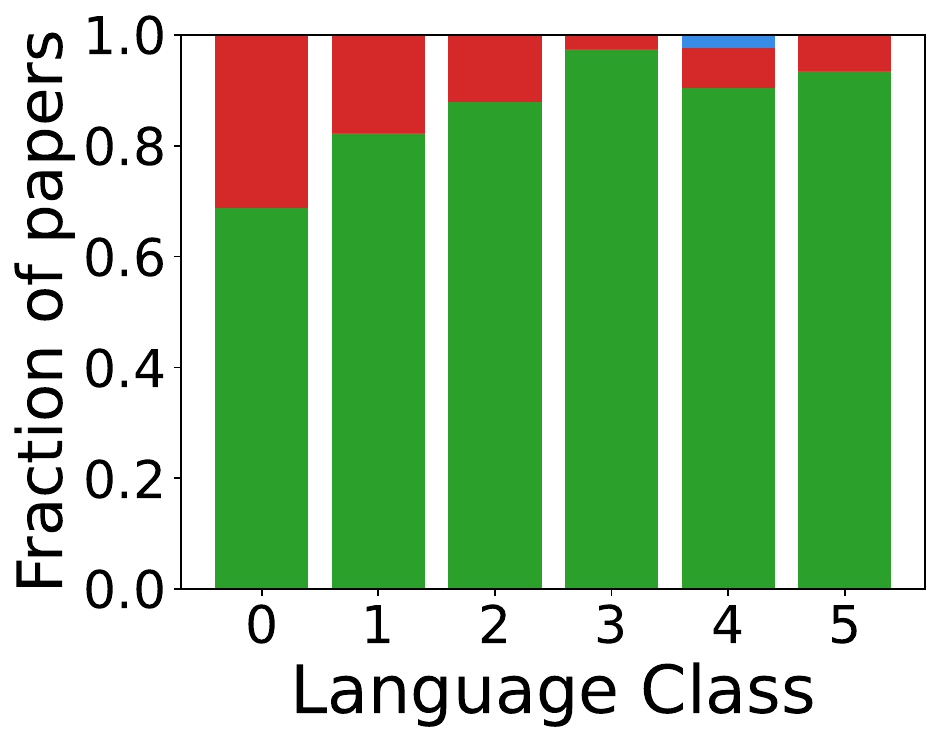}
        \caption{Availability of Code}
        \label{fig:code}
    \end{subfigure}
    \hfill
    \begin{subfigure}{0.24\textwidth}
        \centering
        \includegraphics[width=\linewidth]{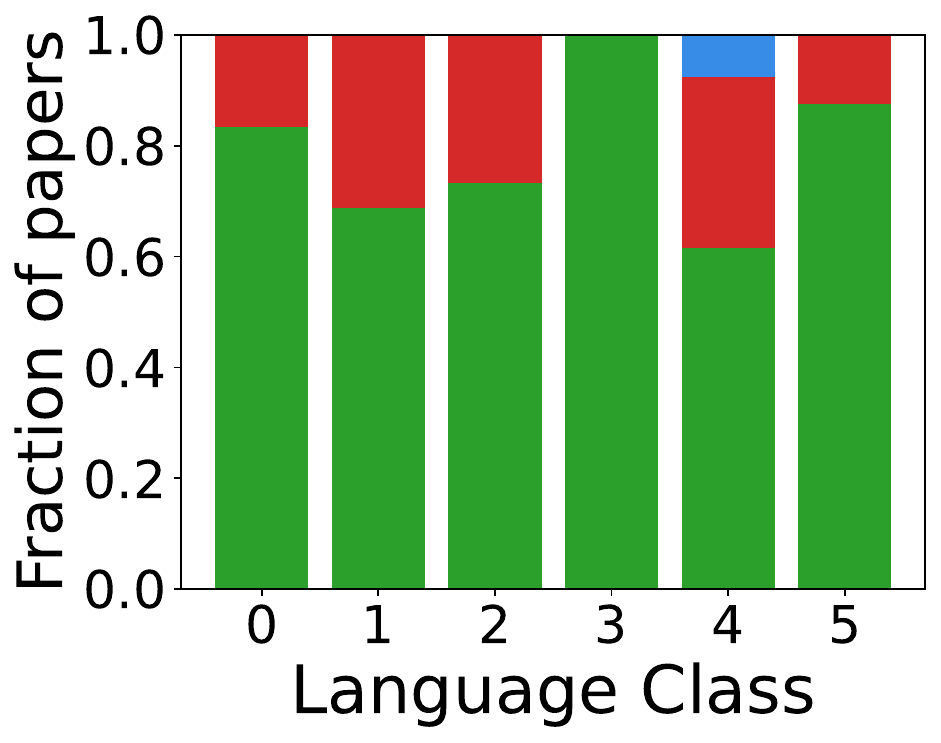}
        \caption{Availability of LM(s)}
        \label{fig:model}
    \end{subfigure}
    \hfill
    \begin{subfigure}{0.24\textwidth}
        \centering
        \includegraphics[width=\linewidth]{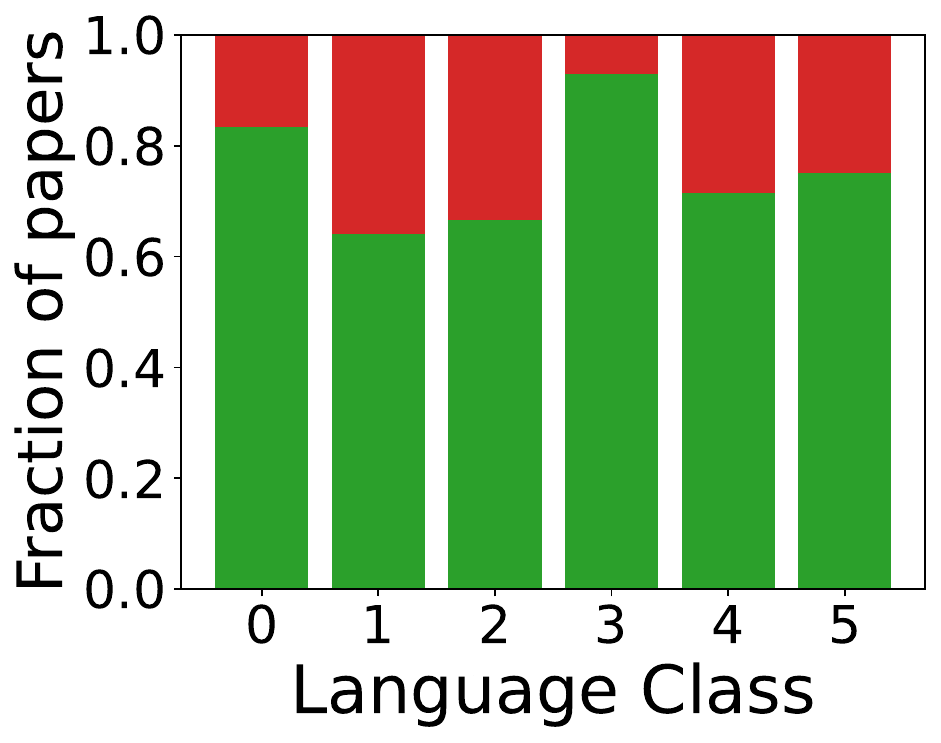}
        \caption{Availability of Tool(s)}
        \label{fig:tool}
    \end{subfigure}
    \centering
    \caption{Analysis on artefact release.}
    \label{fig:resource-availability}
\end{figure*}

We use AA as the research paper repository. While AA is the largest NLP-related paper repository,~\citet{ranathunga-de-silva-2022-languages} note that many papers related to low-resource languages also get published in other venues such as IEEE conferences or regional journals. However, the popularly used Google Scholar does not have a free API to extract data, and the coverage of Semantic Scholar is rather poor\footnote{For example, the search query "english+nlp" returns 4312 results on Semantic Scholar as opposed to the 495,000 results returned by Google Scholar.}. Moreover, some conference and journal publications are hidden behind paywalls. While archives such as arXiv are a possible option, they do not contain the meta data for us to carry out a conference/journal-specific analysis.  Considering all these factors, we selected AA to extract papers for our analysis. AA has been the common choice for many research related to diversity analysis in NLP research~\cite{rungta-etal-2022-geographic, blasi-etal-2022-systematic,cains2019Geographic}.

When collecting data from AA, we reuse data and code from \citet{ranathunga-de-silva-2022-languages} who in turn had used code and data from \citet{blasi-etal-2022-systematic} and \citet{acl_anthology_corpus} (respectively). However, we had to collect data post 2022 by ourselves.

We use the URLs of papers from the ACL Anthology Bibliography to extract the title and abstract of each paper. We then allocate the papers to different languages, following the language list (of {6419} languages) given by~\citet{ranathunga-de-silva-2022-languages}. For each language name, we check for matches in both the title and abstract and download the matched papers using their respective URLs (where a URL to the PDF is available). Of these, 130 languages are ignored due to the high count of false positives caused by matches with existing words and author names\footnote{Examples of languages that were ignored include: \textit{Are}, \textit{As}, \textit{Even}, \textit{One}, \textit{So}, \textit{To}, \textit{Apache}, \textit{U}, \textit{Bit}, \textit{She}.}.
Next, we convert each paper to its text format.

Then we further group these language-wise papers according to language category. The commonly used language category definition that is based on language resources is \citet{joshi-etal-2020-state} (see Table~\ref{tab:language_categories} in Appendix). This definition can be used to categorise languages into six classes, with class 5 being the highest resourced, and class 0 being the least resourced. \citet{joshi-etal-2020-state} used this definition to classify about 2000 languages. However, this categorisation was conducted in 2020 and it has considered only ELRA\footnote{\url{http://www.elra.info/en/}} and LDC\footnote{\url{https://www.ldc.upenn.edu/}} as data repositories. \citet{ranathunga-de-silva-2022-languages} showed that these repositories have very limited coverage for low-resource languages. They reused \citet{joshi-etal-2020-state}'s language category definition and categorised 6419 languages considering the Hugging Face data repository in addition to ELRA and LDC. In this research, we use this newer language categorisation.

\section{Analysis}

\subsection{The degree of artefact reuse in NLP research}
We extract a paper sample of 355 (papers published between 2015-2023) from the dataset downloaded above. To analyse the effect of the publishing venue, these papers are then separated into three categories (henceforth referred to as \textit{PV} categories). These categories are selected based on the suggestion of~\citet{ranathunga-de-silva-2022-languages}.

\begin{itemize}
    \item \textbf{Main}: Main ACL conferences/journals where NLP researchers publish (Full list in Appendix~\ref{sec:appendix_a}).
    \item \textbf{LREC} (Language Resources and Evaluation Conference). It was given a separate category as it is a venue specifically focusing on language resources.
    \item \textbf{Other} - Everything else. Usually, these PVs refer to shared tasks, workshops and regional conferences such as RANLP and ICON.
\end{itemize}

\begin{table}
\centering
\tiny
\begin{tabular}{|l |l|}
\hline
\textbf{Artefact} & \textbf{Status} \\
\hline
\multirow{6}{*}{Data} & Used dataset from some previous research \\
 & Extended an existing dataset \\
 & Used dataset from some previous research but created new data as well \\
 & Introduce new dataset \\
 & Data not needed \\
 & Cannot determine \\
\hline
\end{tabular}
\caption{Possible options for use, and reuse of data}
\label{tab:data_category}
\end{table}

For each PV, the resulting paper sample has 20 papers per language class\footnote{Except for language class 1 in \textit{Main} PV, where we could find only 15 papers.}. We manually read each of these papers to find out whether they created/used data, code\footnote{We considered NLP related tools/libraries/code repositories such as NLTK and Huggingface libraries but did not consider generic libraries such as Pandas.} and/or LMs\footnote{By LMs, we refer to LMs starting from Word2Vec, GloVe and FastText, coming to currently used Large LMs}. The possible options for data-related mentions in a paper are shown in Table~\ref{tab:data_category}. Similar options are considered for code and LMs (see Table~\ref{tab:artefact_des} in Appendix). Note that the first three entries in Tables~\ref{tab:data_category} and \ref{tab:artefact_des} suggest the reuse of artefacts from previous research in some manner.

Out of the {355} papers we analysed, {98.9\%} has reused some form of artefact from previous research. Further language class-wise analysis on this is shown in Figure~\ref{fig:resource-usage-ACL} (In the Appendix we have a larger version in Figure~\ref{fig:resource-usage-ACL-Appendix} as well as a chronological breakdown of the data in Figure~\ref{fig:resource-availability-ACL}).

\textit{Other} PV category is the highest in reusing data as-it-is. This is not surprising, as this category has many papers referring to shared tasks. \textit{Main} category also uses existing data as-it-is to a higher degree, but there is some emphasis on data extension as well. \textit{LREC}, due to its focus on language resources, sees more papers introducing new datasets or extending existing datasets than those that reuse existing data as-it-is. 

The \textit{Main} category sees the highest level of code reuse to introduce new implementations - most papers extend code from already existing research. This has to be due to the highly competitive nature of PVs in this category, where reviewers emphasise technical novelty. \textit{Other} PV category is high in reusing code as well, but it has a relatively higher portion of papers using existing code as-it-is. 

As mentioned earlier, since most \textit{LREC} papers focus on dataset release, they seem not to have paid attention to the use of state-of-the-art solutions involving LMs. In contrast, papers from \textit{Main} heavily emphasise using LMs, and this PV category seems to be the venue to introduce new LMs. 

Overall, the most reused artefact is code, spanning from early APIs/toolkits such as NLTK~\cite{bird2009natural} and Kaldi~\cite{povey2011kaldi} to modern-day Hugging Face libraries. 

\subsection{Percentage of papers that promise to share the newly created artefacts}
Next, we focus on papers that create new artefacts (created from scratch or extended existing artefacts) and report the percentage of papers that promise to share the newly created artefacts. If they do promise, then we check whether they have provided the URL of the public repository containing the artefact(s). 

This analysis was done in a semi-automated manner on the same 355 paper sample as before, using a keyword-based method to filter papers. 

To identify keyword matches, we first replace all non-letter characters of the paper full text with spaces and convert the text to lowercase. To match keywords containing a single term, we split the text by the space character and look for exact matches between the keyword and the words in the resulting array. To match keywords containing multiple terms, we do a direct search over the text (without splitting). We make this distinction between single-word and multi-word keywords due to the false positives caused by matching substrings (for example, "public" would match a text that contains the word "republic"). For each matched keyword, we extract the paragraph in which it was identified and create text files using these paragraphs. These filtered text files assist in identifying the claims of the papers during the manual analysis.

The keywords consist of words that indicate availability. The complete set of keywords is as follows: \texttt{release}, \texttt{released}, \texttt{public}, \texttt{publicly}, \texttt{github}, \texttt{gitlab}, \texttt{huggingface co}, \texttt{osf io}, \texttt{open source}, \texttt{accessible}. Note that the non-letter characters of the keywords are also replaced by spaces to facilitate the matching. Also, note that we do not include keywords such as \texttt{available} and \texttt{http} due to the high number of false positives that they cause. In order to quantify the impact of avoiding these keywords, we look at the false omission rate of a sample of 100 papers. We randomly select 100 papers from the data set and run them through our keyword-based search algorithm. This predicted 69 papers to contain promises of releasing artefacts. We then manually checked the remaining 31 papers in full, to see whether they promised the release of an artefact. Of these 31 papers, one paper has promised and shared the data and code. This results in a false omission rate of approximately 0.03.

We manually read the filtered papers to further verify whether a paper has produced an artefact, and if so, whether it has promised to release that artefact.

Results are shown in Figure~\ref{fig:artefact-release-promises-ACL}. Interestingly, out of the \textit{Main} PV papers that produced some new artefacts, 44\% have not mentioned whether that artefact will be released. In the \textit{Other} category, this value is 67\%. \textit{LREC} has the lowest percentage at 33\%. However, in \textit{LREC}, 36\% of the papers that have promised to release data have not given a repository URL.

\subsection{Further Analysis into Artefact Availability}
\label{sec:further_analysis}
In the above analysis, we can only determine whether a paper mentions that research artefacts are publicly released, and if so, a link to a repository is given. However, that analysis does not tell us the type of these repositories, whether they are accessible, or whether they contain the artefact. Therefore, we carry out a second, more detailed analysis.

To get an insight into more recent trends, we consider papers published between 2020-2023. Following the same semi-automated approach discussed above, we extract a list of papers that promised to release at least one of the following artefacts: \textit{data}, \textit{source code}, \textit{LM}, or \textit{tool}. Then the extracted papers are grouped according to the language class. Classes 5, 4, 3 and 2 have a considerable number of papers, so we sampled 75 from each class. Class 1 and 0 only have 71 and 59 papers, respectively, thus all of those papers were included in our analysis. Altogether, this sample contains 430 papers.

The aggregated result is shown in Figure~\ref{fig:resource-availability}. Be reminded that in this analysis, we omitted the papers that do not refer to an artefact type or those that do not promise to release the artefact they produced. A `No' is marked if a link was not given, a given link is not working, or the repository corresponding to the link does not have the promised artefact (we clicked through and followed all the links mentioned in the papers). 

We notice that a considerable portion of papers that promised to release \textit{data} have `dead-ends' when trying to locate it. This count is higher in low-resource languages. Most tools are hosted on personal or institutional websites, and a portion seems to have fallen out of maintenance in the intervening years. The `dead-end' problem exists to a lesser degree concerning \textit{code} availability. However, even for \textit{code}, class 0 has a noticeable number of `dead-ends'. Overall, most of the links to \textit{code} are active and have the artefact, followed by those that promise to release an \textit{LM}. 

We also record the common repositories used by NLP researchers and provide a summary in Table~\ref{tab:Links} (A breakdown of the same data across language classes is available in Figure~\ref{fig:Links} in the Appendix). According to this, \textit{GitHub} seems to be the most favourite option to release data and code. Some research has considered \textit{Zenodo} and \textit{Hugging Face} for data release\footnote{This result tallies with the survey results published by~\citet{ranathunga-de-silva-2022-languages} to a good extent.}. In contrast, Hugging Face seems to be the favourite choice for LM releases. Most of the tools have their own unique web link, hence the `other' category is the highest for this type.

\begin{table}[!htb]
\centering
\resizebox{\columnwidth}{!}{%
\begin{tabular}{|l|r|r|r|r|r|}
\hline
\textbf{Repository} & \textbf{Code}& \textbf{Data}& \textbf{LMs} & \textbf{Tools}& \textbf{Total} \\
\hline
\hline
GitHub & 153 & 188 & 17 & 12 & 370 \\
Hugging Face & 0 & 6 & 11 & 2 & 19 \\
Zenodo & 1 & 10 & 1 & 0 & 12 \\
Google Drive & 0 & 5 & 3 & 1 & 9 \\
Bitbucket & 4 & 0 & 0 & 1 & 5 \\
GitLab & 3 & 2 & 0 & 0 & 5 \\
Codeberg & 1 & 1 & 0 & 0 & 2 \\
Dropbox & 0 & 1 & 0 & 0 & 1 \\
Mendeley & 0 & 1 & 0 & 0 & 1 \\
Other & 5 & 58 & 6 & 44 & 113 \\
\hline
\textbf{Total} & 167 & 272 & 38 & 60 & 537 \\
\hline
\end{tabular}
}
\caption{Repository usage across all classes}
\label{tab:Links}
\end{table}

\subsection{Analysis Based on NLP Tasks}
Next, we carry out an analysis based on NLP tasks, to understand whether artefact release has any relationship to the type of NLP task\footnote{Initial categorisation of tasks come from Hugging Face task list and a survey paper on NLP research~\cite{de2019survey}}. This analysis was conducted using the paper sample used in Section~\ref{sec:further_analysis}. Table~\ref{tab:data_des} in the Appendix shows the raw counts. \textit{Translation} is the NLP task\footnote{As shown in Table~\ref{tab:data_des}, \textit{Corpora} has the highest raw counts but is not an \textit{NLP Task} per se.} that has the highest number of artefact releases (this artefact is usually parallel data), followed by \textit{morphological analyzer} and \textit{Automatic Speech Recognition (ASR)}. In particular, having morphological analysis as the prevalent NLP domain seems to be common for extremely low-resource languages. This is not surprising - these languages have never had such linguistic resources, and such research is essential in understanding their linguistic properties.  The high amount of ASR-related artefacts could be due to the existence of languages that do not have a writing system\footnote{\citet{ethnologue} notes that around 41\% of the languages they list may be unwritten.}.


\begin{figure}[t]
    \captionsetup[subfigure]{justification=centering}
    \centering
    \begin{subfigure}{0.23\textwidth}  
        \centering
        \includegraphics[width=\linewidth]{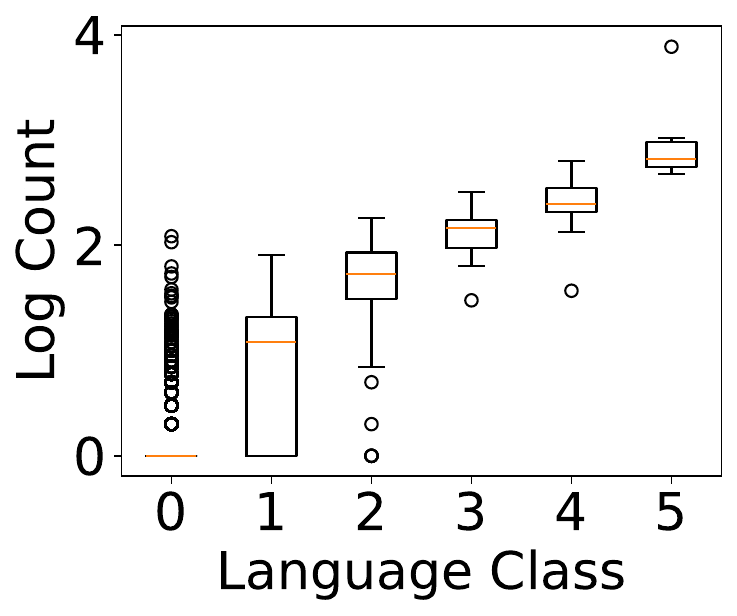}
        \caption{HF Dataset Counts}
        \label{fig:data}
    \end{subfigure}
    \hfill
    \begin{subfigure}{0.23\textwidth}
        \centering
        \includegraphics[width=\linewidth]{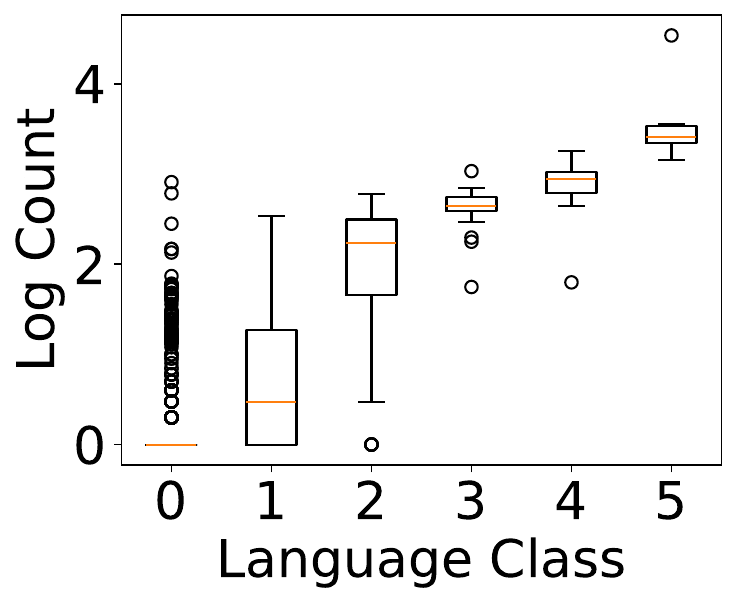}
        \caption{HF LM Counts}
        \label{fig:code}
    \end{subfigure}
    \centering
    \caption{Number of resources for the language classes on Hugging Face (HF).}
    \label{fig:hf-resource}
\end{figure}

\subsection{Dataset and LM Availability}

Our final analysis is based on the datasets and LM counts reported in Hugging Face\footnote{\url{https://huggingface.co/languages}}, which is the fastest-growing repository for NLP-related artefacts. Figure~\ref{fig:hf-resource} shows\footnote{A larger version is available as Figure~\ref{fig:hf-resource-Appendix} in the Appendix.} the language class-wise distribution of data and LMs. 
Further, Table~\ref{tab:num-of-hf} shows relevant numerical values, which demonstrates the language class-wise disparity. 

\begin{table}[!hbt]
    \centering
    \resizebox{\columnwidth}{!}{
    \begin{tabularx}{0.70\textwidth}{|l|Z|Z|Z|Z|Z|Z|}
   \hline 
\multirow{2}{*}{\textbf{Artefact type}}
& \multicolumn{6}{c|}{\textbf{Median of Language Class}}\\
 
 \hhline{~------}
 & \textbf{0} & \textbf{1} & \textbf{2} & \textbf{3} & \textbf{4} & \textbf{5} \\
 \hline
\makecell{Data set\\counts} & 0.0 & 12.0 & 53.0 & 147.5 & 246.0 & 657.0 \\
\hline
\makecell{LM\\ counts} & 0.0 & 3.0 & 171.5 & 443.5 & 881.0 & 2601.0 \\
 \hline
    \end{tabularx}
    }
    \caption{Hugging Face Resource Counts}
    \label{tab:num-of-hf}
\end{table}

The disparity between different language classes is evident from the medians, despite some outliers. Most notably, out of the 6135 languages in class 0, most have no data or LMs, therefore the handful of languages that have some data/LM have become outliers. The correlation between the class-wise LM and data availability is evident - a Pearson correlation value of 0.9972 is reported between the data and LM counts on languages listed in Hugging Face.

\section{Conclusion}
We hope our findings would help the NLP community to better appreciate the benefit of openness and to commit to releasing the artefacts they produce. We further hope these statistics will be useful to ACL in making informed decisions. It would be interesting to run this same experiment 5 or 10 years down the line, to see if there are any changes in releasing and reusing artefacts. In hopes to assist in such efforts, our code is publicly released\footURL{https://bit.ly/ACL2024ShouldersOfGiants}.

\section{Limitations}
We considered only a fraction of the papers published in AA. Our keyword-based paper filtering mechanism might have missed some papers that have made their artefacts available. If a paper does not mention the language name in its abstract, our algorithm does not pick it up. Thus we highly encourage the community to adhere to `Bender Rule'~\cite{bender2019rule}. If a research published their artefact without mentioning that in their paper, or if the link to the artefact was included in a different version of the paper (e.g. ArXiv), such are missed. We might have missed some information on artefacts while manually reading hundreds of research papers, which might have impacted the statistics we present. When checking if a repository link is live, we clicked on that link only once. There could have been instances where the link was momentarily down. In certain instances, we noticed that a URL is not working due to a change in the web repository directory structure. However, we did not try to manually figure out the correct link. We consider an artefact to be available in a repository if we note the availability of files (e.g. python files in a code base) inside the repository. We cannot guarantee the repository has all the artefacts the paper promised (e.g.~all the promised data files or whether the given code is working).

\section{Ethics Statement}
We only used the AA paper repository, which is freely available for research. Our implementation is based on publicly available code. We do not release the paper-wise information we recorded, nor do we re-publish the papers we downloaded from AA.

\bibliography{anthology,custom}

\clearpage
\newpage

\appendix
\section{Language Category Definition}
\label{sec:source}

\begin{table}[!hbt]
\centering
{\small %
\begin{tabularx}{0.49\textwidth}{cXrl}
    \hline
    Class & Description & \multicolumn{2}{c}{Language} \\ 
    \hhline{~~--}
 & & Count & Examples \\
    \hline \hline
    0 & Have exceptionally limited resources, and have rarely been considered in language technologies. & 2191 & \makecell{Slovene\\Sinhala}  \\ \hline
    1 & Have some unlabelled data; however, collecting labelled data is challenging. & 222 & \makecell{Nepali\\Telugu} \\ \hline
    2 & A small set of labelled datasets has been collected, and language support communities are there to support the language. & 19 & \makecell{Zulu\\Irish} \\ \hline 
    3 & Has a strong web presence, and a cultural community that backs it. Have highly benefited from unsupervised pre-training. & 28 & \makecell{Afrikaans\\Urdu} \\ \hline
    4 & Have a large amount of unlabelled data, and lesser, but still a significant amount of labelled data have dedicated NLP communities researching these languages. & 18 & \makecell{Russian\\Ukrainian}\\ \hline
    5 & Have a dominant online presence. There have been massive investments in the development of resources and technologies. & 7 & \makecell{English\\Japanese}\\ \hline
   \end{tabularx}
   }
  \caption{Language Category definition by~\citet{joshi-etal-2020-state}}~\label{tab:language_categories}
  \label{joshi_classes}
\end{table}

\section{Main Conference and Journal List}
\label{sec:appendix_a}
(1) Annual Meeting of the Association for Computational Linguistics, (2) North American Chapter of the Association for Computational Linguistics, (3) European Chapter of the
Association for Computational Linguistics, (4) Empirical Methods in Natural Language Processing,
(5) International Conference on Computational Linguistics, (6) Conference on Computational Natural
Language Learning (7) International Workshop on
Semantic Evaluation, (8) Conference of the Asia-Pacific Chapter of the Association for Computational Linguistics, and (9) Conference on Computational Natural Language Learning.

In addition, the following journals are considered: (1) Transactions of the Association for Computational Linguistics and (2) Computational
Linguistics. 

\section{Artefact Annotation Scheme}

All the annotators involved in this study are coauthors of the paper. In Table~\ref{tab:artefact_des} we show the annotation scheme we used. 

\begin{table}[!htb]
\centering
\tiny
\begin{tabularx}{0.5\textwidth}{|l|X|}
\hline
\textbf{Artefact} & \textbf{Status} \\
\hline
\hline
\multirow{6}{*}{Data}  & Used dataset from some previous research \\
\hhline{~-}
 & Extended an existing dataset \\
 \hhline{~-}
 & Used dataset from some previous research but created new data as well \\
 \hhline{~-}
 & Introduce new dataset \\
 \hhline{~-}
 & Data not needed \\
 \hhline{~-}
 & Cannot determine \\
 \hline
 \hline
\multirow{6}{*}{Code} & Used an implementation from some previous research \\
\hhline{~-}
 & Extended an existing implementation (e.g.~toolkit, library) \\
 \hhline{~-}
 & Used an implementation from some previous research but implemented part of the solution from scratch\\
 \hhline{~-}
 & Provided their implementation \\
 \hhline{~-}
 & Code not needed \\
 \hhline{~-}
 & Cannot determine \\
 \hline
 \hline
\multirow{6}{*}{LM} & Used an existing LM \\
\hhline{~-}
 & Extended an existing LM\\
 \hhline{~-}
 & Used an existing LM but trained their LM(s) as well\\
 \hhline{~-}
 & Trained their own LM \\
 \hhline{~-}
 & LM not needed \\
 \hhline{~-}
 & Cannot determine \\
\hline
\end{tabularx}
\caption{Possible options for Artefacts}
\label{tab:artefact_des}
\end{table}

\section{Code and Data Reuse}
Code and data from \citet{ranathunga-de-silva-2022-languages} and \citet{acl_anthology_corpus} are released under \textbf{CC BY-NC 4.0} licence. The authors obtained permission from~\citet{blasi-etal-2022-systematic} to use the code on their public repository\footURL{https://github.com/neubig/globalutility}. 

\section{NLP Task Breakdown Across Language Classes}

We show the NLP task breakdown across the five language classes in Table~\ref{tab:data_des}.

\begin{table*}
\centering
\resizebox{\textwidth}{!}{%
\begin{tabularx}{\textwidth}{|l|Z|Z|Z|Z|Z|Z|Z|Z|Z|}
\hline
\multirow{2}{*}{\textbf{NLP Task}} & \multicolumn{6}{c|}{Language Class}  & \multirow{2}{*}{\textbf{Total}}\\
\hhline{~------~}
& \textbf{0}& \textbf{1}& \textbf{2} & \textbf{3}& \textbf{4} &\textbf{5} &\\
\hline
\hline
Corpora & 19 & 22 & 11 & 11 & 11 & 29 & 103 \\
Translation & 10 & 12 & 8 & 6 & 10 & 6 & 52 \\
Morphological Analyzer & 11 & 8 & 2 & 3 & 1 & 0 & 25 \\
Automatic Speech Recognition (ASR) & 5 & 1 & 10 & 4 & 3 & 0 & 23 \\
Language Model & 1 & 2 & 10 & 1 & 2 & 5 & 21 \\
Parsers & 4 & 5 & 3 & 1 & 3 & 4 & 20 \\
Data Sets & 6 & 1 & 5 & 3 & 4 & 0 & 19 \\
Dictionary/Lexicon & 6 & 4 & 1 & 1 & 3 & 3 & 18 \\
Named-Entity Recognition (NER) & 1 & 0 & 3 & 5 & 7 & 2 & 18 \\
Text Classification & 1 & 2 & 1 & 2 & 1 & 9 & 16 \\
Part of Speech (PoS) & 1 & 6 & 3 & 2 & 2 & 1 & 15 \\
Cross-Lingual Applications & 2 & 1 & 3 & 6 & 2 & 0 & 14 \\
Text Generation & 0 & 0 & 0 & 0 & 6 & 4 & 10 \\
Hate Speech Detection & 0 & 0 & 2 & 6 & 1 & 0 & 9 \\
Misinformation Detection & 0 & 0 & 0 & 4 & 3 & 1 & 8 \\
Wordnets/Ontology/Taxonomy & 3 & 1 & 0 & 2 & 0 & 1 & 7 \\
Discourse Analysis & 0 & 2 & 1 & 1 & 2 & 1 & 7 \\
Question and Answer (QnA) & 0 & 1 & 2 & 1 & 3 & 0 & 7 \\
NLP Tools & 1 & 4 & 1 & 0 & 0 & 0 & 6 \\
Semantic (Other) & 0 & 0 & 0 & 0 & 1 & 5 & 6 \\
Tokenizer & 0 & 0 & 1 & 2 & 0 & 2 & 5 \\
Semantic Similarity & 0 & 0 & 0 & 3 & 1 & 1 & 5 \\
Multiple Tasks & 0 & 0 & 1 & 3 & 0 & 0 & 4 \\
Spelling and Grammar & 0 & 1 & 1 & 0 & 1 & 1 & 4 \\
Summarizing & 0 & 0 & 0 & 3 & 1 & 0 & 4 \\
Phonological Analyzer & 0 & 1 & 0 & 2 & 1 & 0 & 4 \\
Sentiment Analyzer & 0 & 0 & 1 & 1 & 2 & 0 & 4 \\
Text-to-Speech & 0 & 2 & 1 & 0 & 0 & 0 & 3 \\
Transliteration & 0 & 0 & 2 & 0 & 1 & 0 & 3 \\
Lexical Inference & 0 & 0 & 0 & 0 & 3 & 0 & 3 \\
Coreference Resolution & 0 & 0 & 0 & 0 & 3 & 0 & 3 \\
Information Extraction & 0 & 0 & 1 & 1 & 0 & 0 & 2 \\
Bilingual Lexicon Induction (BLI) & 0 & 1 & 0 & 0 & 1 & 0 & 2 \\
Optical Character Recognition (OCR) & 0 & 1 & 0 & 0 & 0 & 1 & 2 \\
Language Identification (LangID) & 0 & 0 & 1 & 0 & 0 & 0 & 1 \\
Intent Detection & 1 & 0 & 0 & 0 & 0 & 0 & 1 \\
News/Social Media Recommendation & 0 & 0 & 0 & 0 & 1 & 0 & 1 \\
Text Classification & 0 & 0 & 1 & 0 & 0 & 0 & 1 \\
Stemming & 0 & 0 & 0 & 0 & 1 & 0 & 1 \\
\hline
\textbf{Total} & 72 & 78 & 76 & 74 & 81 & 76 & 457 \\
\hline
\end{tabularx}
}
\caption{NLP Tasks Conducted}
\label{tab:data_des}
\end{table*}

\begin{figure*}[!hbt]
    \captionsetup[subfigure]{justification=centering}
    \begin{subfigure}{0.99\textwidth}  
        \centering
        \includegraphics[width=\linewidth]{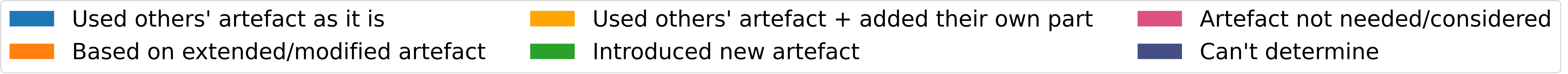}
    \end{subfigure}
    
    \centering
    \begin{subfigure}{0.32\textwidth} 
        \centering
        \includegraphics[width=\linewidth]{figures/classwise_resource_usage_flipped/classwise_resource_usage_LREC-data.pdf}
        \caption{Usage of Data - LREC}
        \label{fig:LRECdata-Appendix}
    \end{subfigure}
    \hfill
    \begin{subfigure}{0.32\textwidth}
        \centering
        \includegraphics[width=\linewidth]{figures/classwise_resource_usage_flipped/classwise_resource_usage_LREC-code.pdf}
        \caption{Usage of Code - LREC}
        \label{fig:LRECcode-Appendix}
    \end{subfigure}
    \hfill
    \begin{subfigure}{0.32\textwidth}
        \centering
        \includegraphics[width=\linewidth]{figures/classwise_resource_usage_flipped/classwise_resource_usage_LREC-model.pdf}
        \caption{Usage of LMs - LREC}
        \label{fig:LRECmodel-Appendix}
    \end{subfigure}

    \begin{subfigure}{0.32\textwidth}  
        \centering
        \includegraphics[width=\linewidth]{figures/classwise_resource_usage_flipped/classwise_resource_usage_Main-data.pdf}
        \caption{Usage of Data - Main}
        \label{fig:Maindata-Appendix}
    \end{subfigure}
    \hfill
    \begin{subfigure}{0.32\textwidth}
        \centering
        \includegraphics[width=\linewidth]{figures/classwise_resource_usage_flipped/classwise_resource_usage_Main-code.pdf}
        \caption{Usage of Code - Main}
        \label{fig:Maincode-Appendix}
    \end{subfigure}
    \hfill
    \begin{subfigure}{0.32\textwidth}
        \centering
        \includegraphics[width=\linewidth]{figures/classwise_resource_usage_flipped/classwise_resource_usage_Main-model.pdf}
        \caption{Usage of LMs - Main}
        \label{fig:Mainmodel-Appendix}
    \end{subfigure}   

    \begin{subfigure}{0.32\textwidth}  
        \centering
        \includegraphics[width=\linewidth]{figures/classwise_resource_usage_flipped/classwise_resource_usage_Other-data.pdf}
        \caption{Usage of Data - Other}
        \label{fig:Otherdata-Appendix}
    \end{subfigure}
    \hfill
    \begin{subfigure}{0.32\textwidth}
        \centering
        \includegraphics[width=\linewidth]{figures/classwise_resource_usage_flipped/classwise_resource_usage_Other-code.pdf}
        \caption{Usage of Code - Other}
        \label{fig:Othercode-Appendix}
    \end{subfigure}
    \hfill
    \begin{subfigure}{0.32\textwidth}
        \centering
        \includegraphics[width=\linewidth]{figures/classwise_resource_usage_flipped/classwise_resource_usage_Other-model.pdf}
        \caption{Usage of LMs - Other}
        \label{fig:Othermodel-Appendix}
    \end{subfigure}  
    \centering
    \caption{Artefact (Data, Code, LM) creation, extension, and reuse across ACL venues - Aggregated analysis}
    \label{fig:resource-usage-ACL-Appendix}
\end{figure*}

\section{Code and Data Intended Use}
All the code use was consistent with their intended use as specified on the relevant research publications~\citep{ranathunga-de-silva-2022-languages,blasi-etal-2022-systematic} and the \verb|readme| files on the repositories~\citep{acl_anthology_corpus}.

\section{Artefact Creation, Extension, and Reuse}

\begin{figure*}[!hbt]
    \begin{subfigure}{0.99\textwidth}  
        \centering
        \includegraphics[width=\linewidth]{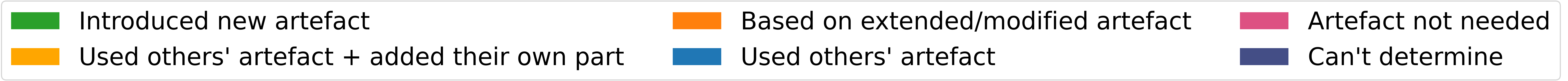}
    \end{subfigure}
    
    \centering
    \begin{subfigure}{0.32\textwidth} 
        \centering
        \includegraphics[width=\linewidth]{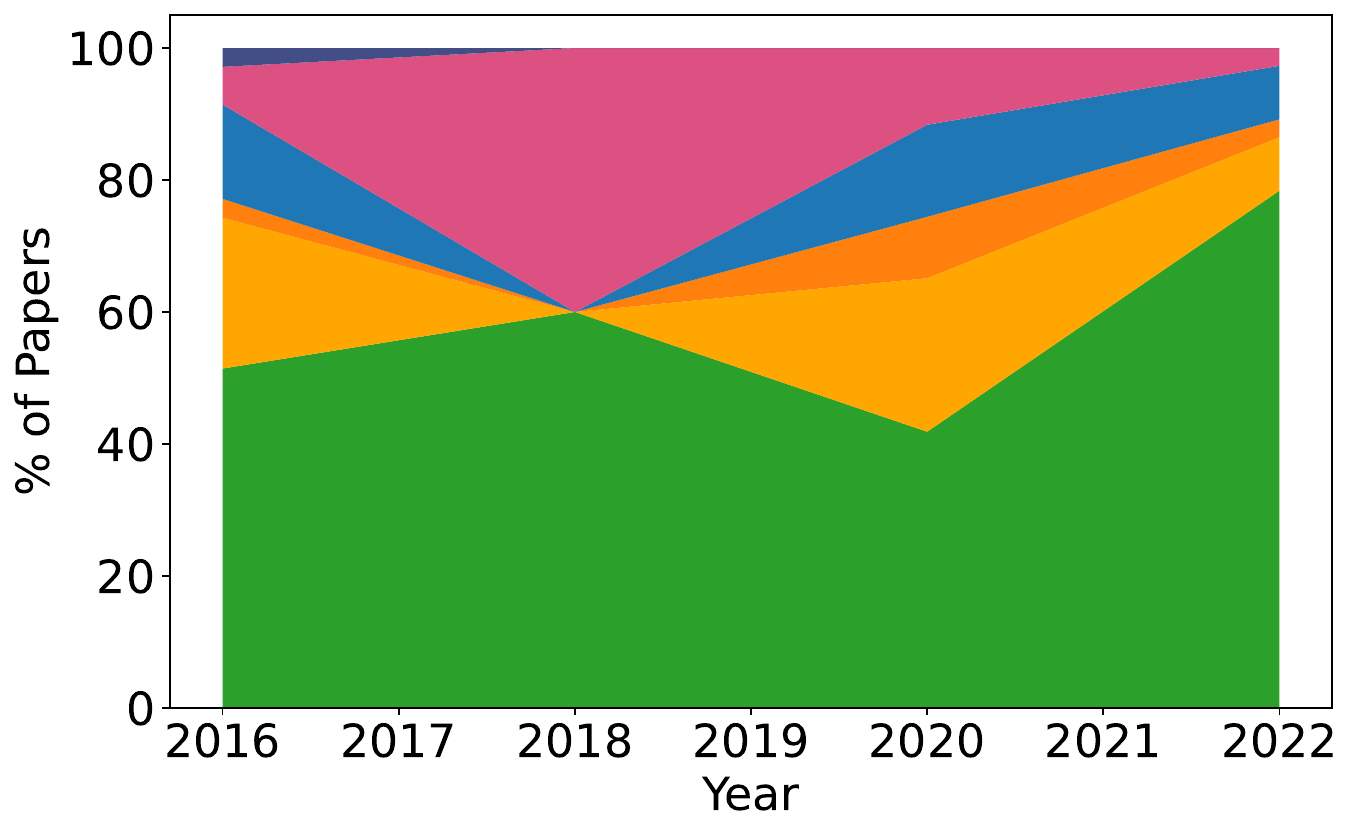}
        \caption{Usage of Data - LREC}
        \label{fig:LRECdata}
    \end{subfigure}
    \hfill
    \begin{subfigure}{0.32\textwidth}
        \centering
        \includegraphics[width=\linewidth]{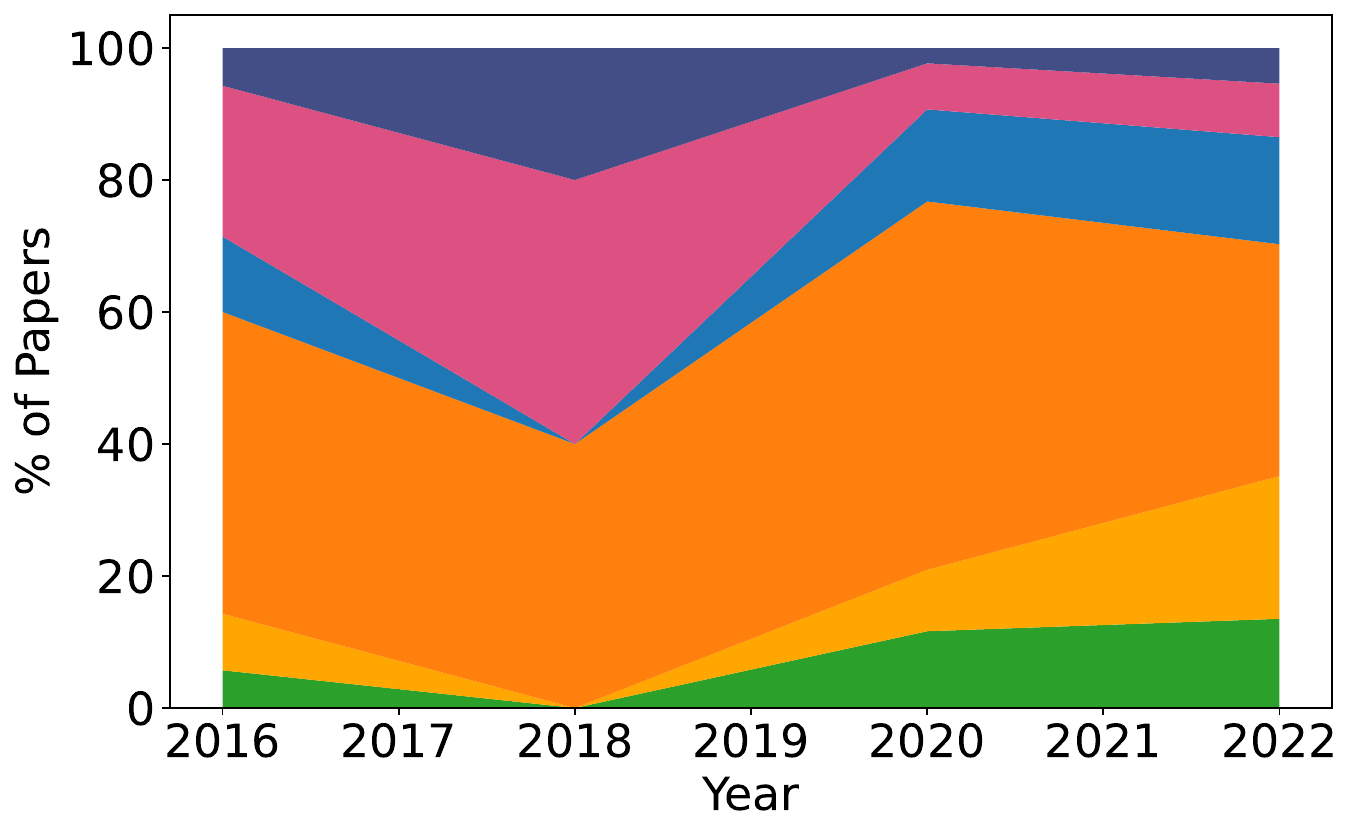}
        \caption{Usage of Code - LREC}
        \label{fig:LRECcode}
    \end{subfigure}
    \hfill
    \begin{subfigure}{0.32\textwidth}
        \centering
        \includegraphics[width=\linewidth]{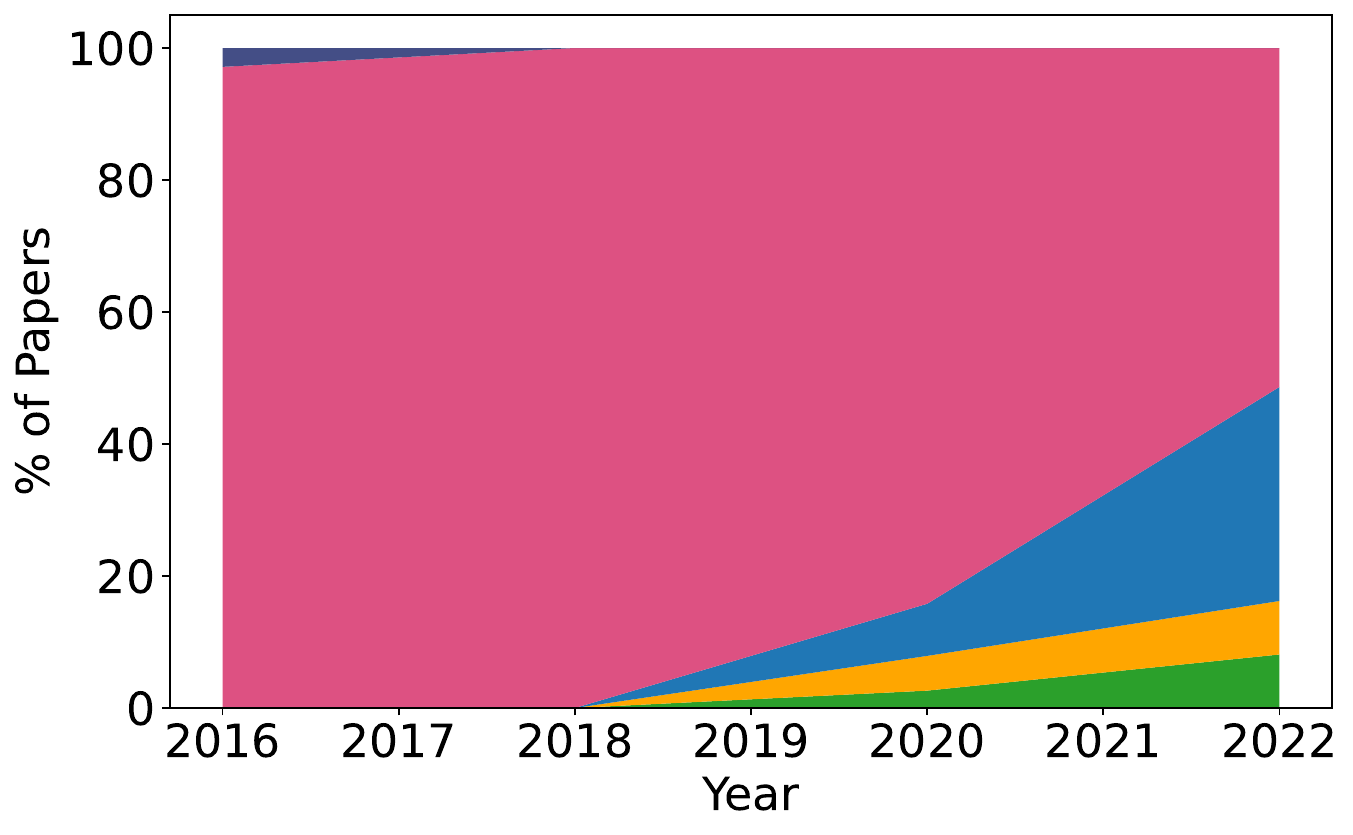}
        \caption{Usage of LMs - LREC}
        \label{fig:LRECmodel}
    \end{subfigure}

    \begin{subfigure}{0.32\textwidth}  
        \centering
        \includegraphics[width=\linewidth]{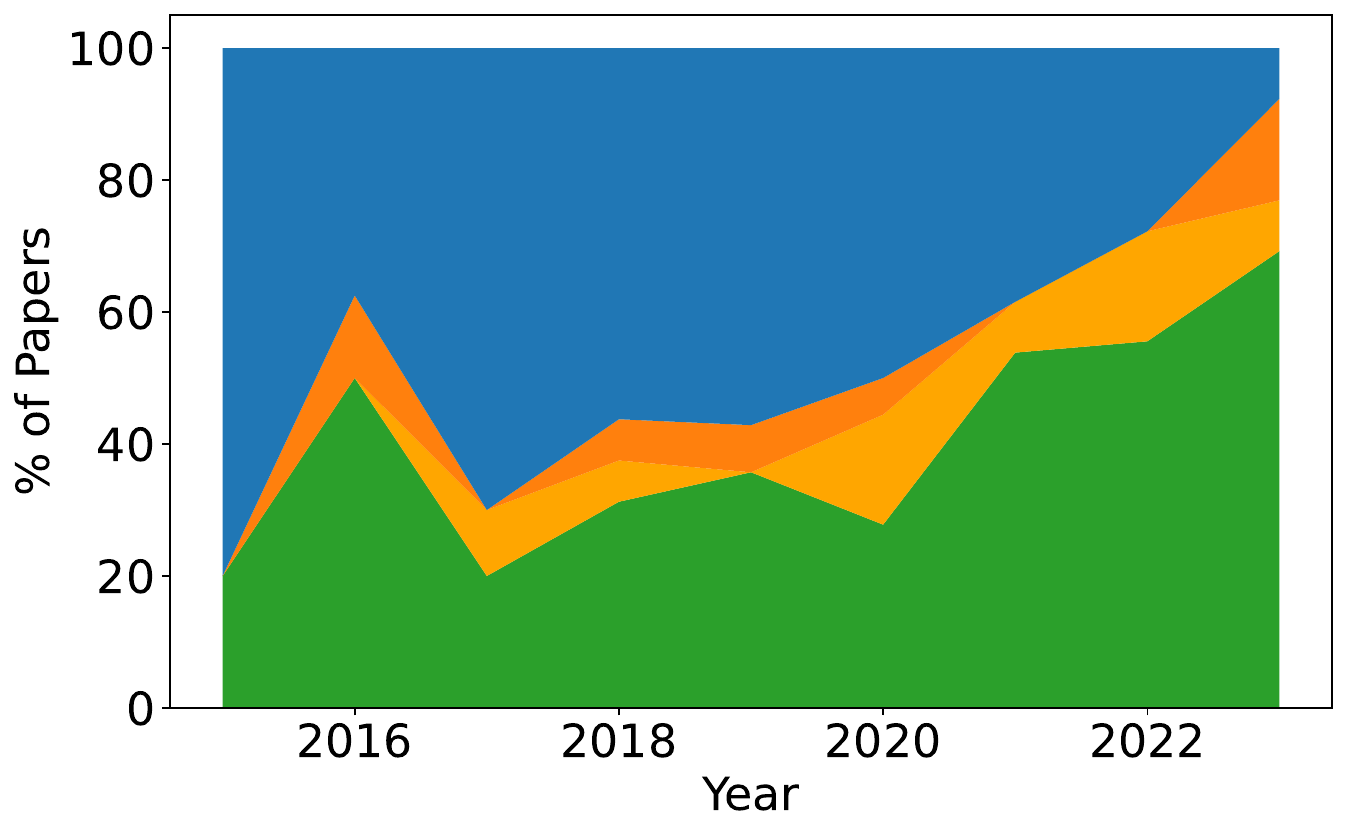}
        \caption{Usage of Data - Main}
        \label{fig:MainData}
    \end{subfigure}
    \hfill
    \begin{subfigure}{0.32\textwidth}
        \centering
        \includegraphics[width=\linewidth]{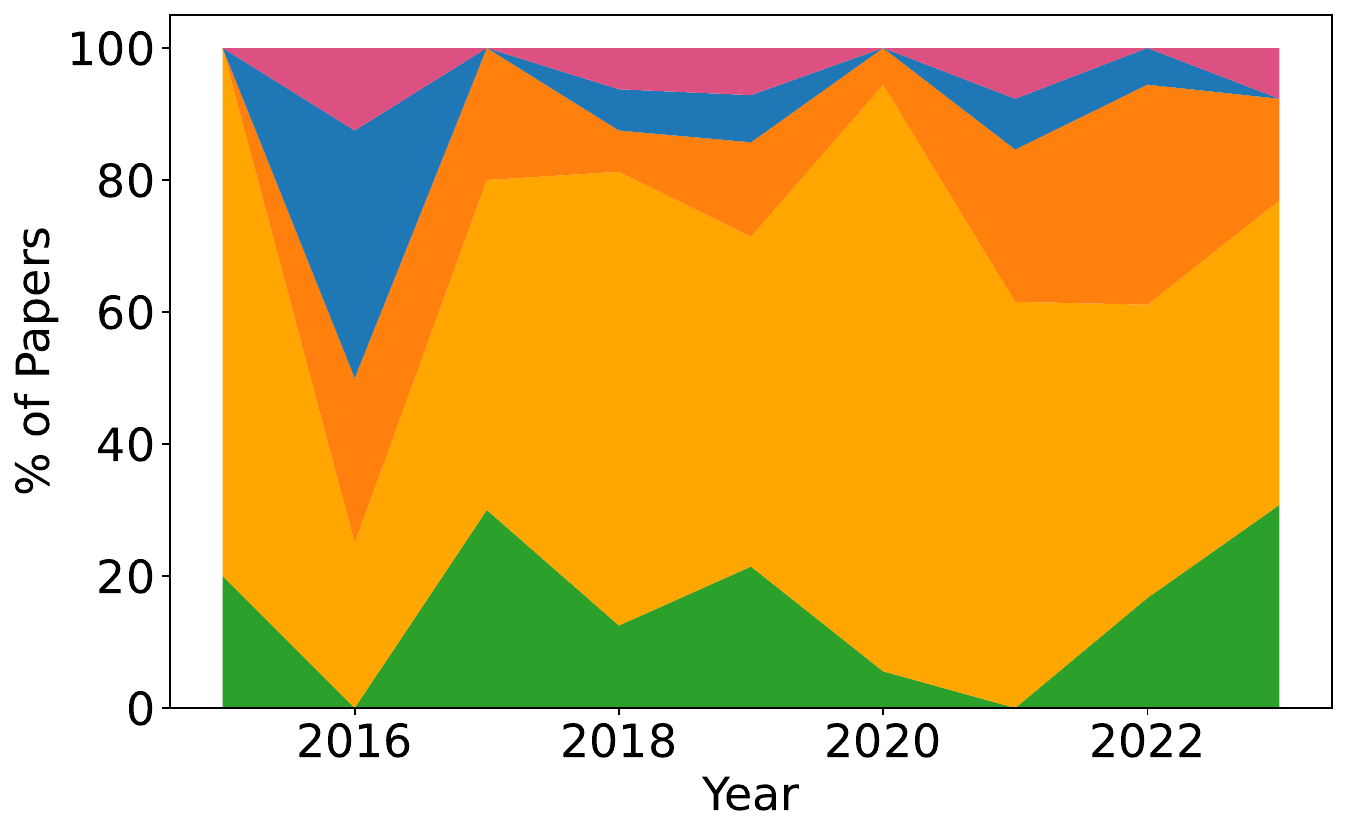}
        \caption{Usage of Code - Main}
        \label{fig:MainCode}
    \end{subfigure}
    \hfill
    \begin{subfigure}{0.32\textwidth}
        \centering
        \includegraphics[width=\linewidth]{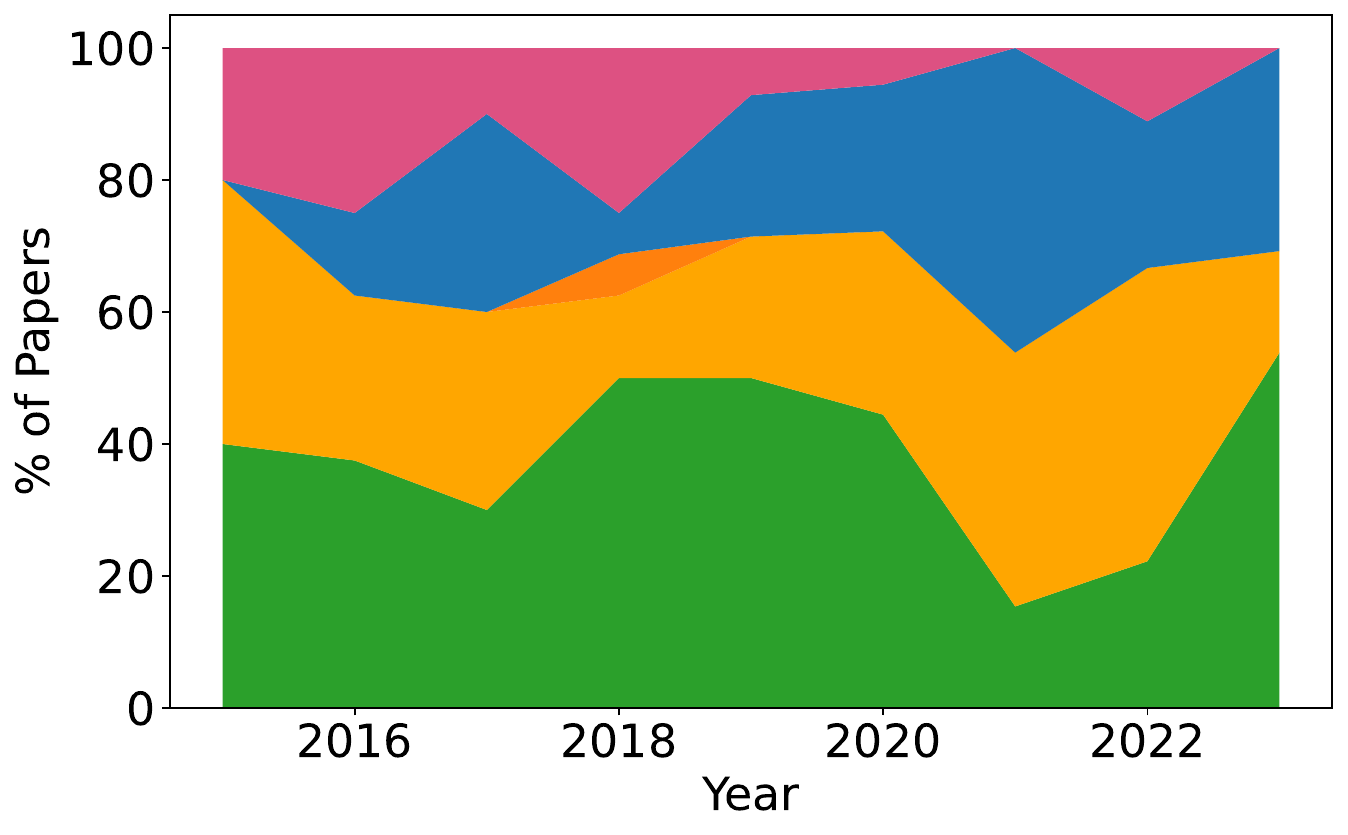}
        \caption{Usage of LMs - Main}
        \label{fig:MainModel}
    \end{subfigure}   

    \begin{subfigure}{0.32\textwidth}  
        \centering
        \includegraphics[width=\linewidth]{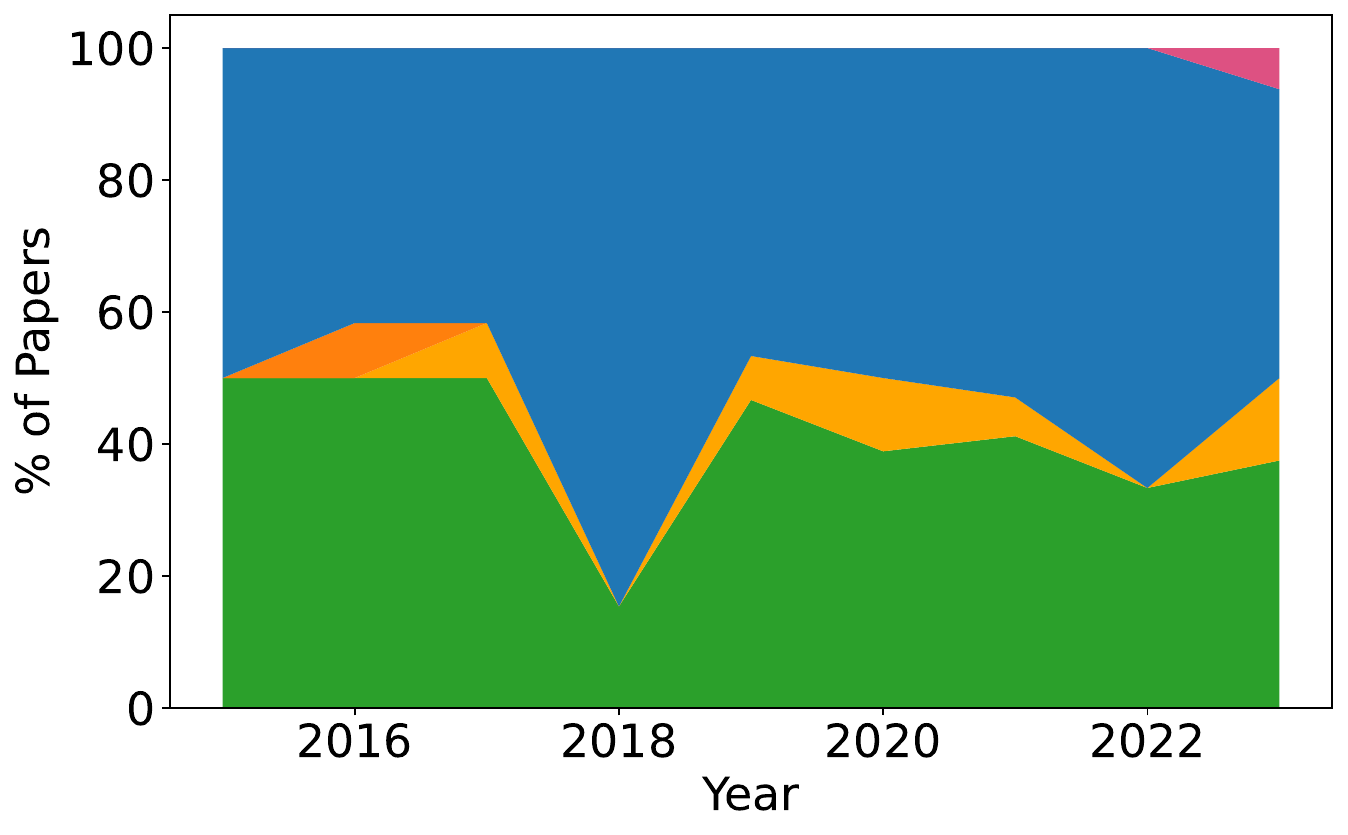}
        \caption{Usage of Data - Other}
        \label{fig:OtherData}
    \end{subfigure}
    \hfill
    \begin{subfigure}{0.32\textwidth}
        \centering
        \includegraphics[width=\linewidth]{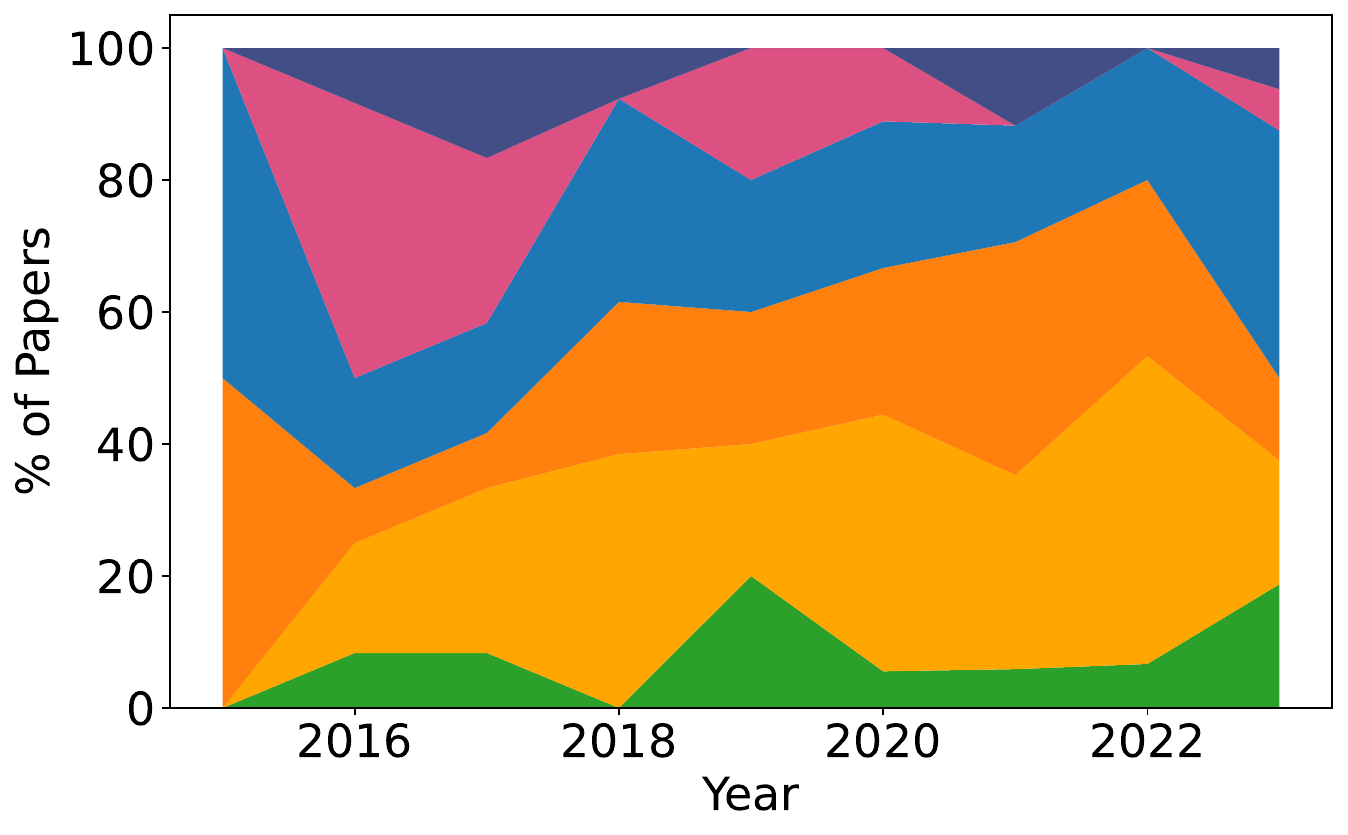}
        \caption{Usage of Code - Other}
        \label{fig:OtherCode}
    \end{subfigure}
    \hfill
    \begin{subfigure}{0.32\textwidth}
        \centering
        \includegraphics[width=\linewidth]{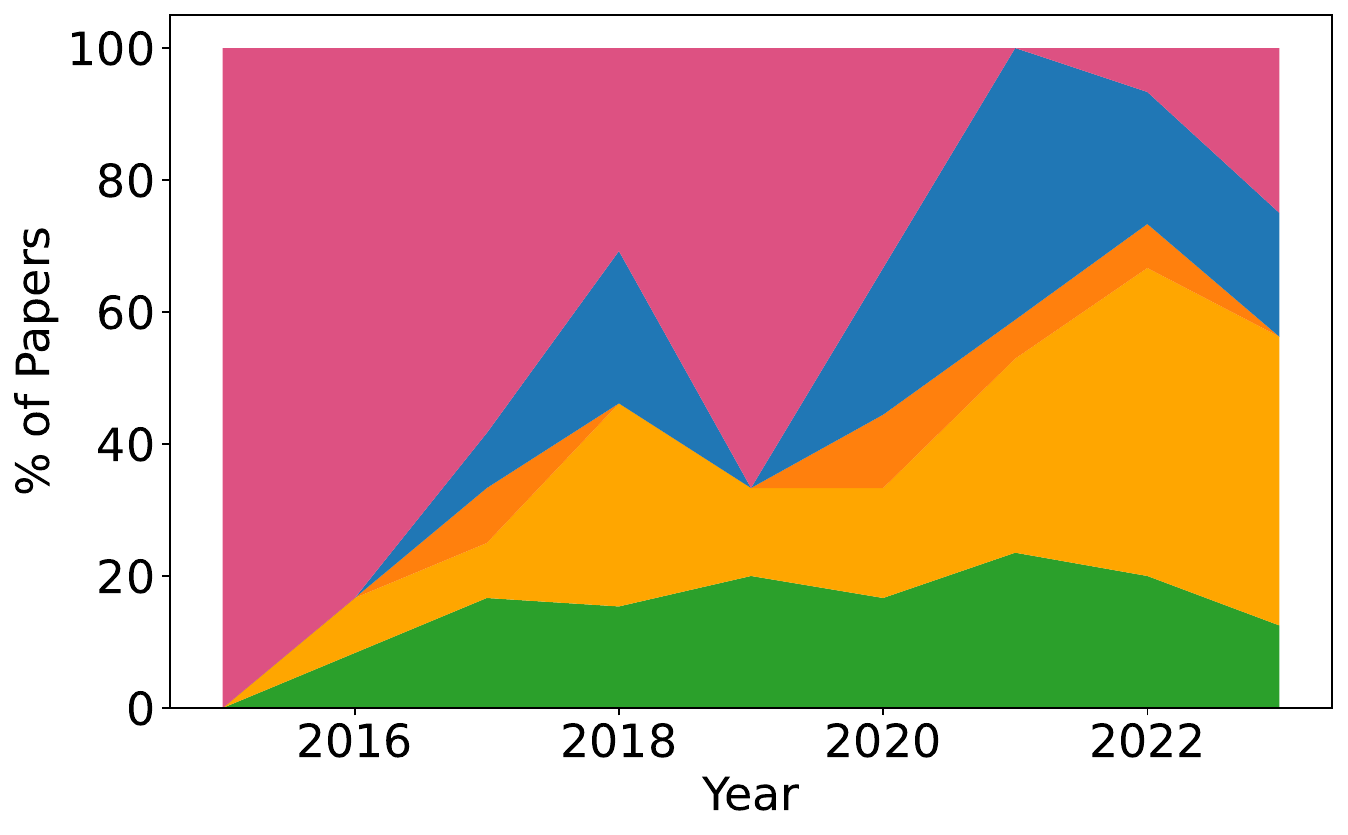}
        \caption{Usage of LMs - Other}
        \label{fig:OtherModel}
    \end{subfigure}  
    \centering
    \caption{Cumulative percentage graphs - Artefact (Data, Code, LM) creation, extension, and reuse across ACL venues. - Chronological analysis.}
    \label{fig:resource-availability-ACL}
\end{figure*}

In Figure~\ref{fig:resource-usage-ACL-Appendix} we have the larger version of the Figure~\ref{fig:resource-usage-ACL} for improved readability. Further, given that the information in Figure~\ref{fig:resource-usage-ACL-Appendix} is presented after aggregating across time but separated into language classes, we also include a set of cumulative percentage graphs in Figure~\ref{fig:resource-availability-ACL} where we show the same data aggregated across the language classes but spread out over the publication years to better show the changing trends in resource availability and reuse.
Unsurprisingly, as per Figures~\ref{fig:LRECcode}, \ref{fig:MainCode}, and \ref{fig:OtherCode}, we can see that \textit{code} is being re-used the most across all venues. LREC (Figure~\ref{fig:LRECdata}) stands out among the \textit{data} graphs (Figures~\ref{fig:MainData} and \ref{fig:OtherData}) for consistently being a source of new data sets rather than a venue where existing data is reused. We see that LMs, had a reasonable presence in the main venues (Figure~\ref{fig:MainModel}) even before our analysis period while in the \textit{other} venues (Figure~\ref{fig:OtherModel}), the trend stars just at the beginning of our considered time period. LREC on the other hand, seems to be late to be considered for LMs as it is only in 2018, that we see them becoming noticeable in Figure~\ref{fig:LRECmodel}.

\section{Artefact Hosting}
\label{sec:artefact_host}

Table~\ref{tab:Links} shows a summary of where NLP researchers have published their data, based on the information mentioned in the research papers. According to this, \textit{GitHub} seems to be the most favourite option to release data and code. Some research has considered \textit{Zenodo} and \textit{Hugging Face} for data release\footnote{This result tallies with the survey results published by~\citet{ranathunga-de-silva-2022-languages} to a good extent.}. In contrast, Hugging Face seems to be the favourite choice for LM releases. Most of the tools have their own unique web link, hence the `other' category is the highest for this type.

\begin{figure}[!hbt]
    \captionsetup[subfigure]{justification=centering}
    \centering
    \begin{subfigure}{0.46\textwidth}  
        \centering
        \includegraphics[width=\linewidth]{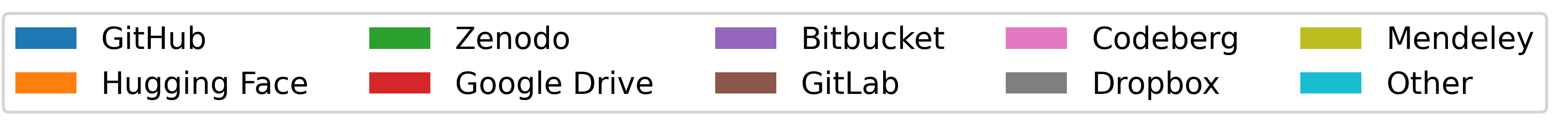}
    \end{subfigure}
    
    \centering
    \begin{subfigure}{0.23\textwidth}  
        \centering
        \includegraphics[width=\linewidth]{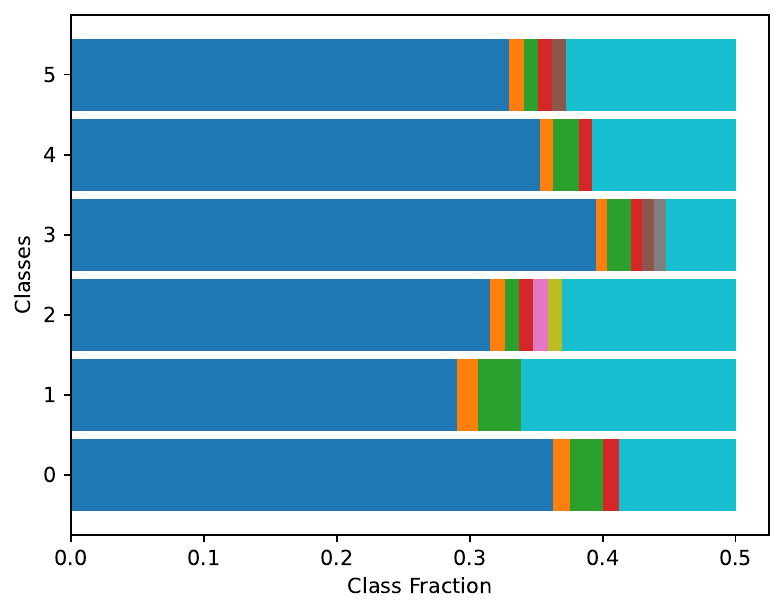}
        \caption{Data}
        \label{fig:Linksdata}
    \end{subfigure}
    \hfill
    \begin{subfigure}{0.23\textwidth}
        \centering
        \includegraphics[width=\linewidth]{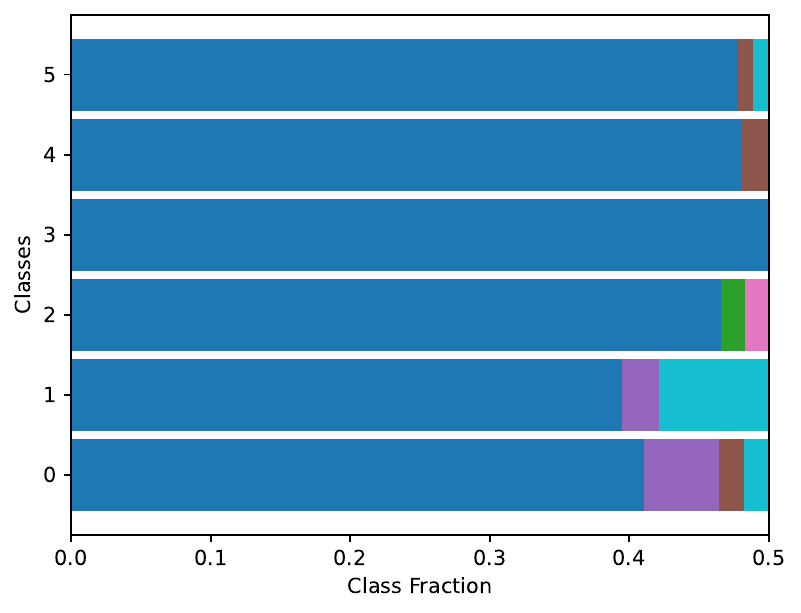}
        \caption{Code}
        \label{fig:Linkscode}
    \end{subfigure}  
    
    \begin{subfigure}{0.23\textwidth}  
        \centering
        \includegraphics[width=\linewidth]{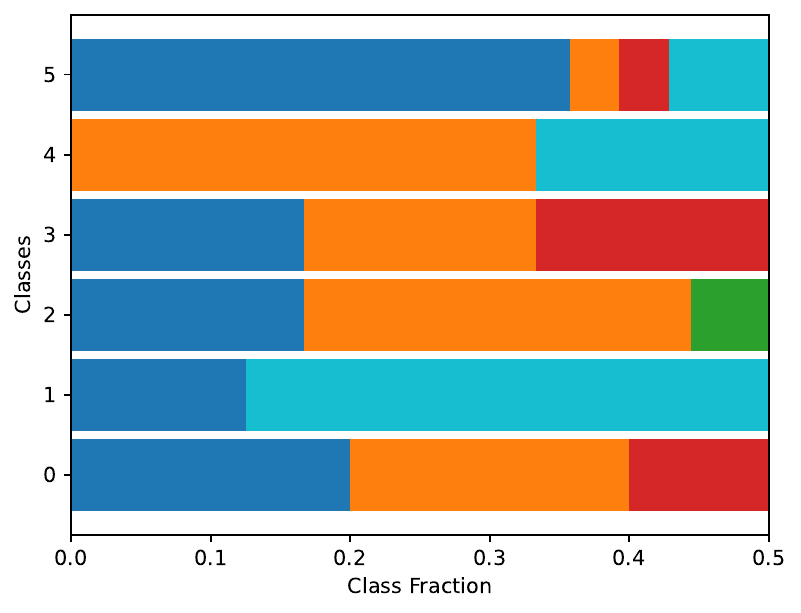}
        \caption{LMs}
        \label{fig:LinksModels}
    \end{subfigure}
    \hfill
    \begin{subfigure}{0.23\textwidth}
        \centering
        \includegraphics[width=\linewidth]{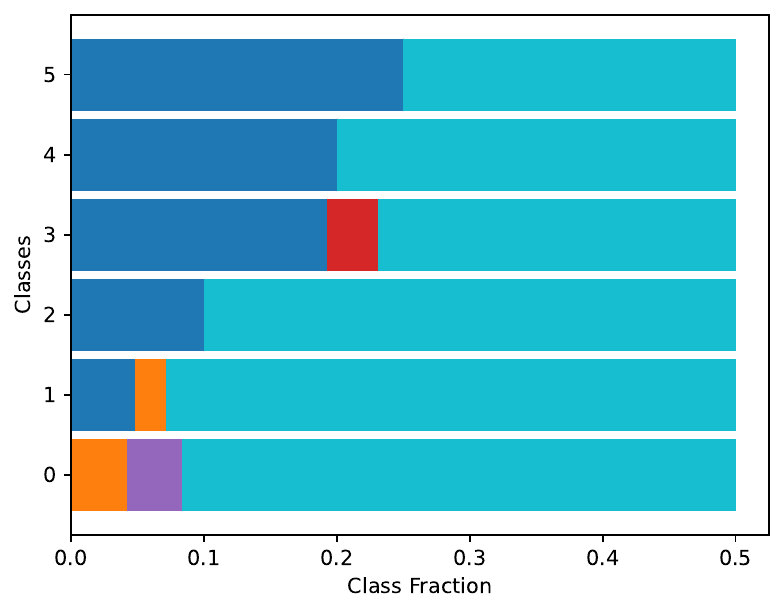}
        \caption{Tools}\label{fig:LinksTools}
    \end{subfigure}  
    \centering
    \caption{Artefact (Data, Code, LM, Tools) hosting locations.}
    \label{fig:Links}
\end{figure}

In Figure~\ref{fig:Links} we show a more detailed view of the artefacts being hosted online; previously discussed in Table~\ref{tab:Links} as a summary. Here it is possible to note the variations between the language classes. For example, the interesting observation of Figure~\ref{fig:LinksModels} is that it can be noted that while researchers in all other listed language classes use github to host their trained LMs, the researchers of Class 4 languages opt for Hugging Face. Conversely, from Figure~\ref{fig:LinksTools}, it can be noted that in Class 0 languages, tools are generally not hosted on github.
A curious observation in Figure~\ref{fig:LinksModels} is that for some reason, Class 1 languages do not select Hugging Face as a clear contender to host their language models, something that all other language classes seem to do. The overwhelming prevalence of the \textit{other} option in Figure~\ref{fig:LinksTools} can be explained by the fact that most tools tend to be hosted on dedicated websites. Even when the actual site is hosted on a service such as github, they are masked with shorter and more market-friendly custom URLs. 

\section{Hugging Face Resources}

\begin{figure*}[!hbt]
    \captionsetup[subfigure]{justification=centering}
    \centering
    \begin{subfigure}{0.48\textwidth}  
        \centering
        \includegraphics[width=\linewidth]{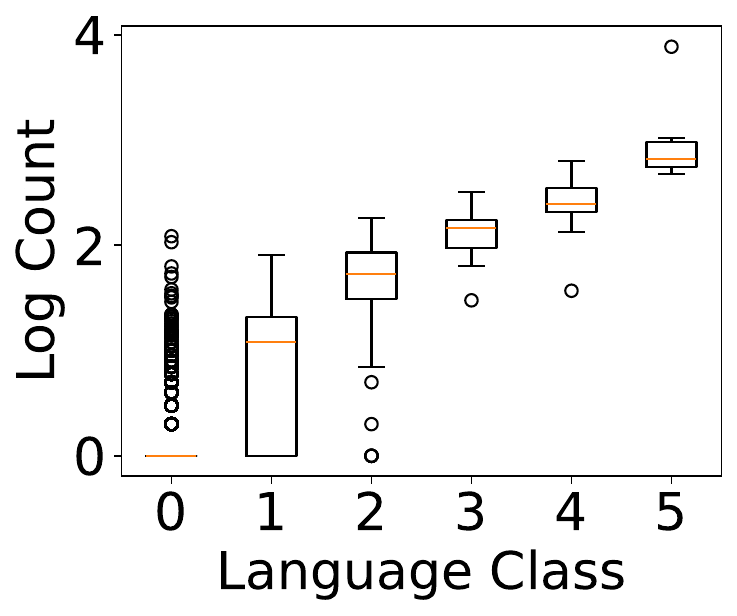}
        \caption{HF Dataset Counts}
        \label{fig:data-Appendix}
    \end{subfigure}
    \hfill
    \begin{subfigure}{0.48\textwidth}
        \centering
        \includegraphics[width=\linewidth]{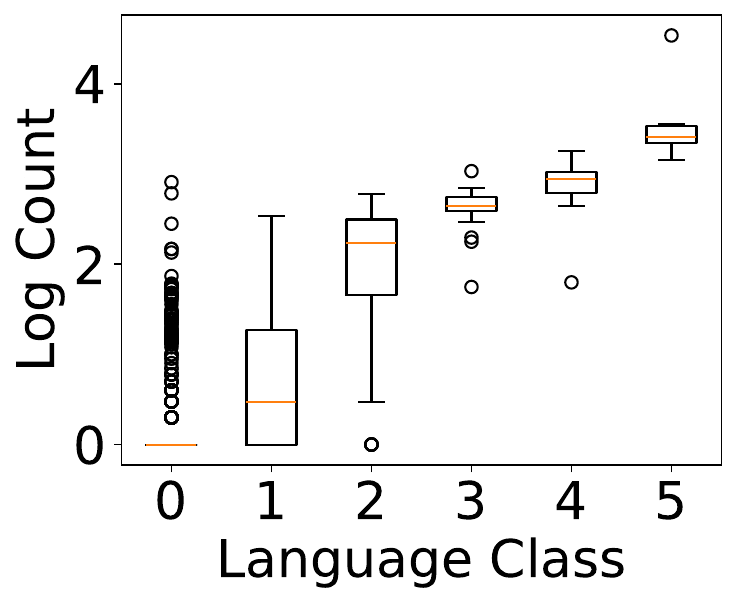}
        \caption{HF LM Counts}
        \label{fig:code-Appendix}
    \end{subfigure}
    \centering
    \caption{Number of Hugging Face (HF) resources for the language classes.}
    \label{fig:hf-resource-Appendix}
\end{figure*}

In Figure~\ref{fig:hf-resource-Appendix} we show the resources available on Hugging Face for the 5 language classes. This is a larger version of the Figure~\ref{fig:hf-resource} for improved readability. Note especially how the entire interquartile range of class 0 is at zero due to the dearth of resources existing for the languages in that class. Thus a language in class 0 with \textit{any} amount of resources gets registered as an outlier. On the opposite end of the spectrum, note class 5 with only 7 languages in the set even after the reclassification by~\citet{ranathunga-de-silva-2022-languages}. Despite that, English still manages to be an outlier with its exceptional resource availability. 

From Figure~\ref{fig:hf-resource-Appendix} and Table~\ref{tab:num-of-hf}, it can be observed a considerable jump between the median values when comparing adjacent classes. This may be taken as both: 1) an indication of the visible difference in the resource availability of the language classes, 2) A reaffirmation of the soundness of the class borders proposed by by~\citet{ranathunga-de-silva-2022-languages} as the distinct medians can be taken as a quality of classes which are internally cohesive and mutually separate.   


\end{document}